\pdfoutput=1

\documentclass[11pt]{article}

\usepackage[]{acl}

\usepackage{times}
\usepackage{latexsym}
\usepackage{blindtext}
\usepackage{graphicx}
\usepackage{multirow} 
\usepackage{multicol} 
\usepackage{float}
\usepackage{booktabs}
\usepackage{svg}
\usepackage{pdfpages}

\usepackage{enumitem}
\usepackage{geometry}
\usepackage{lipsum}
\usepackage{graphicx}
\usepackage[T1]{fontenc}
\usepackage{subcaption}
\usepackage{comment}

\usepackage[utf8]{inputenc}

\usepackage{microtype}

\newcommand{\aakanksha}[1]{\textbf{\small {\color{blue}[#1 -Aakanksha]}}}

\title{On-the-fly Definition Augmentation of LLMs for Biomedical NER}

\author{Monica Munnangi$^\diamondsuit$* \quad Sergey Feldman$^\clubsuit$ \quad  \textbf{Byron C Wallace}$^{\diamondsuit}$ \quad \\
\textbf{Silvio Amir}$^\diamondsuit$ \quad \textbf{Tom Hope}$^\clubsuit$ \quad \textbf{Aakanksha Naik}$^\clubsuit$
  \\
$^\diamondsuit$Northeastern University \\
$^\clubsuit$ Allen Institute for AI \\
\texttt{\small \{munnangi.m,b.wallace,s.amir\}@northeastern.edu} \\
\texttt{\small \{aakankshan,sergeyf,tomh\}@allenai.org}}

\begin{document}
\maketitle
\begingroup\def\thefootnote{*}\footnotetext{ Work perfomed during internship at AI2}\endgroup
\begin{abstract}

Despite their general capabilities, LLMs still struggle on biomedical
NER tasks, which are difficult due to the presence of specialized terminology and lack of training data. 
In this work we set out to improve LLM performance on biomedical NER in limited data settings via a new knowledge augmentation approach which incorporates definitions of relevant concepts on-the-fly. 
During this process, to provide a test bed for knowledge augmentation, we perform a comprehensive exploration of prompting strategies. 
Our experiments show that definition augmentation is useful for both open source and closed LLMs.
For example, it leads to a relative improvement of 15\% (on average) in GPT-4 performance (F1) across all (six) of our test datasets. 
We conduct extensive ablations and analyses to demonstrate that our performance improvements stem from adding relevant definitional knowledge. 
We find that careful prompting strategies also improve LLM performance, allowing them to outperform fine-tuned language models in few-shot settings.  
To facilitate future research in this direction, we release our code at \url{https://github.com/allenai/beacon}.



\end{abstract}
\section{Introduction}
\label{section:intro}

Despite the impressive zero- and few-shot capabilities of LLMs generally, their performance on named entity recognition (NER) over biomedical text remains underwhelming. 
For instance, \citet{gutiérrez2022thinking} observe that using GPT-3 \cite{brown2020language} with \emph{in-context learning} 
performs worse than a smaller, fine-tuned pretrained language model given the same amount of data. 
Despite significant real-world utility, several aspects make this task challenging even for state-of-the-art LLMs. Biomedical texts 
use specialized terminology that often requires domain expertise to interpret. 
In addition to complicating the task, this requirement of requisite background knowledge makes annotation expensive, time-consuming, and difficult to acquire, resulting in limited availability of labeled data. 

\begin{figure}[!t]
\centering
  \includegraphics[scale=0.7]{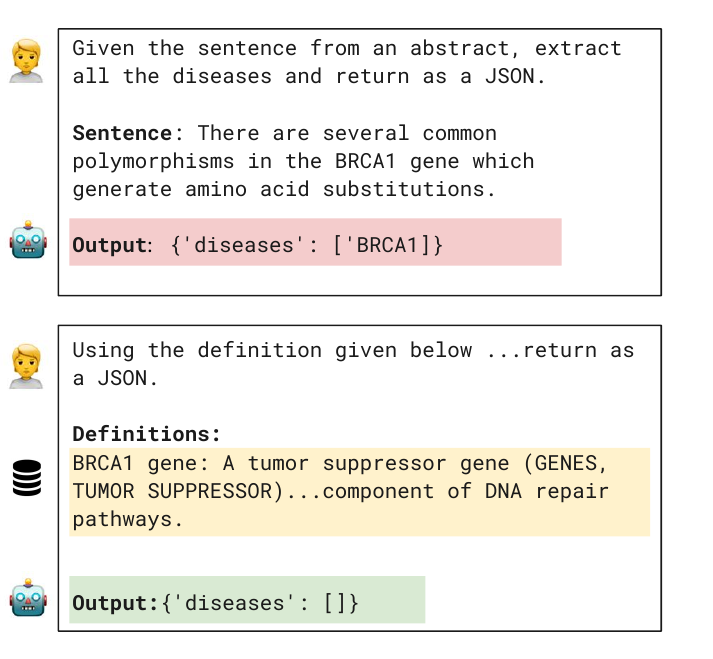}
  \caption{Illustration of our approach using a zero-shot
example, with incorrect extraction (red) and correct extraction (green) when provided with the definition of the extracted entity (yellow).}
  \label{fig:example}
\end{figure}

LLMs have shown promising improvements in performance on general information extraction (IE) tasks \cite{Ashok2023PromptNERPF,wadhwa2023revisiting}. 
Motivated by this, we aim to improve their performance on a specific domain (biomedicine) via a new knowledge augmentation approach which incorporates definitions of relevant concepts dynamically.  
To facilitate this, we perform a comprehensive exploration of prompting strategies; this provides a solid test bed for experimenting with knowledge augmentation for NER. 
More specifically, we first design an experimental framework for assessment of LLMs on biomedical NER (\S~\ref{sec:expframework}). Starting from the BigBIO \cite{fries2022bigbio} collection of 100+ biomedical datasets, we systematically construct an evaluation 
set consisting of six NER datasets. These cover extraction tasks of varying complexity, ranging from open extraction (i.e., no entity types) to extraction according to large, fine-grained schemas (10+ types).
We use this test bed to benchmark the performance of a series of SOTA LLMs, both open and closed, on biomedical NER in both zero-shot and few-shot settings (\S~\ref{sec:ICLforNER}). 

Our benchmarking effort includes an extensive exploration of 
prompting strategies which have provided utility in recent work on using LLMs for IE such as using definitions/explanations \cite{Ashok2023PromptNERPF} and producing extractions in structured formats like code \cite{dunn2022structured,li2023codeie}. 
To the best of our knowledge, this is the first effort investigating such methods for biomedical NER, and we report promising results. 
In particular, we find that these strategies enable LLMs to surpass smaller, fine-tuned LMs in few-shot settings, contrary to prior work.

Building on these strong baselines, we propose a knowledge augmentation approach to further improve LLM performance. 
Our approach, illustrated in Figure~\ref{fig:example}, focuses on identifying and providing definitions of relevant biomedical concepts as a \emph{follow-up} step at inference time, allowing the model to correct its entity extractions. 

We explore two strategies for follow-up prompting: (i) Single-turn, which requires models to make all entity corrections in a single step, and; (ii) Iterative prompting, which simplifies the correction task by allowing models to make changes one entity at a time. 
Our results show that definition augmentation provides meaningful performance improvements across the LLMs considered (including both closed and open models). 
For example, including definitions increases GPT-4 performance by 15\% on average across the datasets we use for evaluation.  

Through a series of ablations, we verify that these performance improvements are due to the presence of relevant concept definitions.  
For example, we find that adding irrelevant definitional knowledge 
yields little to no performance improvement. 
Finally, we evaluate the utility of definitions retrieved from various human-curated sources (UMLS, WikiData) as well as ones automatically generated using LLMs, and find that human-curated definitions lead to higher performance improvements. 
Our results raise interesting questions about the value of definitional knowledge for improving LLM performance on different tasks and across diverse domains where data is limited.  

\section{Experimental Framework}
\label{sec:expframework}


\textbf{Models} We evaluate SOTA LLMs over a set of biomedical NER datasets from the BigBio benchmark~\cite{fries2022bigbio}. 
We assess a variety of models including closed models available via API---i.e., Open AI's GPT-3.5~\cite{brown2020language} and GPT-4~\cite{openai2023gpt4} and Anthropic's Claude 2~\cite{Claude2}---and an open-source model (Llama 2; ~\citealt{touvron2023LLaMA}). 
We list all models in Table~\ref{tab:models}. 
We also conducted preliminary experiments with Google's PaLM~\cite{chowdhery2022palm} but found its performance subpar and so did not pursue further. 

\vspace{0.25em}
\noindent \textbf{Evaluation} We evaluate all models according to entity-level F1. 
Prior work has shown that strict F1 may underestimate the performance of generative models on information extraction tasks, because such models can generate outputs that differ from reference  annotations but which are still correct~\cite{wadhwa2023revisiting}. 
To address this, we complement our automatic evaluation with manual evaluation of a subset of examples; we present this in Appendix  ~\ref{ssec:error}.

\begin{table}[ht!]
\centering
\small
\begin{tabular}{p{0.45\linewidth}p{0.25\linewidth}clp{0.35\linewidth}}
\toprule
\textbf{Dataset} & \textbf{Entity Types}  & \textbf{Size} \\
\midrule

    \textbf{CHEM} \cite{krallinger2017overview} & Chemicals, Proteins & 800  \\

    \textbf{CDR} \cite{DBLP:journals/biodb/LiSJSWLDMWL16} & Chemicals, Diseases & 500 \\

    \textbf{NCBI} \cite{dougan2014ncbi}  & Diseases  & 100 \\
    \midrule
    
    \textbf{MEDM} \cite{mohan2019medmentions} & Biomedical Concepts & 879 \\

    \textbf{PICO} \cite{nye-etal-2018-corpus} & Populations, Interventions, Outcomes & 187 \\


    \textbf{CHIA} \cite{kury2020chia} & Clinical Trial Criteria & 600  \\
    
\bottomrule
\end{tabular}
\caption{Overview of all datasets included in our final biomedical NER evaluation test bed. The size column reports the size of the test split.}
\label{tab:datasets}
\end{table}

\noindent
\textbf{Dataset Selection} As a test bed for biomedical NER, we select datasets from the BigBIO benchmark, a meta-resource of 100+ datasets sourced from various areas of biomedicine, covering 12 task types and 10+ languages. 
NER is the dominant task category in BigBIO, consisting of 76 datasets~\cite{fries2022bigbio}. 
We narrow these down by first excluding datasets that contain: Clinical/EHR data, social media content, and non-English texts.
Several of the remaining datasets contain annotations for the same entity types. 
Therefore, we further filter the corpora by retaining only 1-2 \emph{representative} datasets for all entity types. 
This filtering yields 16 datasets. We further narrow the selection of datasets based on two factors: (i) Prevalence in common benchmarks such as three datasets (CDR, CHEM, NCBI) of the final six, are included in popular biomedical benchmarks (BLURB, BLUE, BoX, and so on) or; (ii) Presence of interesting IE phenomena that could be challenging for LLMs, such as the presence of long entities (PICO), large fine-grained entity type schema (CHIA) and open-ended entity extraction (MEDM).
These datasets are summarized in Table~\ref{tab:datasets} and further described in Table~\ref{tab:datasetexamples}, which also provides examples.

\section{ICL for Biomedical NER}
\label{sec:ICLforNER}

In this section we establish the baseline performance of LLMs in zero- and few-shot settings over all datasets. 
To contextualize these results, we also report on the performance of a smaller, fine-tuned model (Flan-T5 XL; \citealt{chung2022scaling}).

\subsection{Zero-Shot Experimental Setup}
\label{ssec:zeroshot}


We evaluate zero-shot prompting strategies along two main axes: (i) Input format, controls how the task description and expected target categories are provided to the model; (ii) Output format, controls how the model structures outputs.

We explore two possible types of input format: (i) \textbf{Text}, using a standard prompt with a brief description of the task and a list of valid target entity types to be extracted; and (ii) \textbf{Schema Def}, augmenting the standard prompt with detailed descriptions of all target entity types following prior work \cite{Ashok2023PromptNERPF, shao2023prompt}.




For output format, we explore two types of \emph{structured} formats: (i) \textbf{JSON}~\cite{dunn2022structured,li2023evaluating}, and (ii) \textbf{Code} snippets~\cite{li2023codeie,wang2023code4struct}. Recent work has shown that such formats improve zero-shot IE performance of LLMs, while producing valid extractions which are easier to post-process and evaluate.

Our zero-shot experiments evaluate the performance of all four combinations of input and output formats on all models (except GPT-4, omitted in these experiments given the high costs of querying the API). Example prompts for each combination are presented in Appendix~\ref{apx:prompts}.

\begin{table*}[ht]
\centering
\small
\begin{tabular}{lcccccccc} 
\toprule
\textbf{Model}& \textbf{Input} & \textbf{Output}& \textbf{CHEM} & \textbf{CDR} & \textbf{MEDM} & \textbf{NCBI} & \textbf{PICO} & \textbf{CHIA}\\
\midrule
\multirow{ 4}{*}{\textbf{GPT-3.5}}  & Text & JSON & 49.60 & 65.64 & \textbf{43.42} & \textbf{54.05} & 10.71 & 7.43 \\ 
   &  & Code & ~ 42.31&	50.72	&42.91	&44.23	&14.88	& \textbf{31.28} \\ \cmidrule{2-9}
 & + Schema Def & JSON & 47.70 & 64.74 & \textbf{43.72} & 46.79 & 9.53 & 4.72 \\ 
 &  & Code & 41.49&	51.16&	42.46	&47.13	&13.52&	29.43 \\
\midrule
\multirow{ 2}{*}{\textbf{Claude 2}} 
  & Text & JSON & 56.36 & \textbf{67.96} & 36.39 & 44.17 & 7.70 & 19.96  \\ 
   &  +Schema Def & JSON & 45.19 & 60.51 & 34.30 & 37.93 & 4.81 & 19.11 \\
\midrule
\multirow{ 4}{*}{\textbf{Llama 2}} 
  & Text & JSON & \textbf{59.75} &	66.77 &	28.93	& 34.23	& 7.49 &	4.03\\ 
   &  & Code & 57.53	& 55.18	& 23.69& 	24.64	& \textbf{15.39}	& 21.59  \\ \cmidrule{2-9}
 & +Schema Def & JSON & 52.47	& 55.47	& 23.05	& 28.22 & 	3.95	& 3.32\\ 
 &  & Code & 56.04 & 	54.91	& 28.82	& 24.05	& 15.12	& 7.49 \\
\bottomrule
\end{tabular}
\caption{Zero-shot scores with \emph{text input and JSON output},  \emph{text input and code output}, 
\emph{definition input and JSON output} and 
\emph{definition input and code output}, with an exception of Claude 2 which we experimented on JSON (did not output executable code). }
\label{tab:zeroshot}
\end{table*}

\begin{table*}[ht!]
\centering
\small
\resizebox{!}{3.2cm}{
\begin{tabular}{lccccccc} 
\toprule
\textbf{Model} & \textbf{\#Shots} & \textbf{CHEM} & \textbf{CDR} & \textbf{MEDM} & \textbf{NCBI} & \textbf{PICO} & \textbf{CHIA}\\
\midrule
        
\multirow{ 3}{*}{\textbf{GPT-3.5}} & 0 & 49.60 & 65.64 & 43.42 & 54.05 & 14.88 & 31.28 \\
 & 1& 56.06 ($\pm$ 1.03) & 64.05 ($\pm$ 2.92) & 49.15 ($\pm$ 1.69) & 44.27 ($\pm$ 2.59) & 15.83 ($\pm$ 1.9) &	33.72 ($\pm$0.99) \\
 & 3 &  59.54 ($\pm$ 2.24) & 67.44 ($\pm$ 0.52) & 48.47 ($\pm$ 1.63) & 54.20 ($\pm$ 1.53) & 17.11 ($\pm$ 1.65)	& 34.8 ($\pm$0.65)\\
 & 5& 58.66 ($\pm$ 0.79) & 68.19 ($\pm$ 1.07) & 48.10 ($\pm$ 1.28) & 56.02 ($\pm$ 1.48) & 17.12 ($\pm$3.83)	&36.47 ($\pm$0.6) \\
\midrule
\multirow{ 3}{*}{\textbf{Claude 2}} 
& 0& 56.36 & 67.96 & 36.39 & 44.17 & 7.70 & 19.96\\
& 1& 55.19 ($\pm$ 2.21) & 66.43 ($\pm$ 3.08) & 44.82 ($\pm$ 3.04) & 37.89 ($\pm$ 13.42) & 6.3 ($\pm$ 1.2) & 18.94 ($\pm$ 1.43) \\
& 3 & 59.68 ($\pm$ 1.61) & 68.13 ($\pm$ 6.01) & 48.20 ($\pm$ 1.91) & 43.89 ($\pm$ 1.63) & 6.21  ($\pm$ 2.6) &	19.87 ($\pm$ 3.41) \\
& 5& 63.04 ($\pm$ 0.21) & 69.74 ($\pm$ 1.47) & 48.12 ($\pm$ 1.45) & 42.99 ($\pm$ 1.59) & 6.12 ($\pm$ (8.21) & 19.88 ($\pm$ 1.63) \\
\midrule
\multirow{ 3}{*}{\textbf{Llama 2}} 
& 0& 59.75 & 66.77 & 28.93 & 34.23 & 15.39 & 21.59\\
& 1&  57.11 ($\pm$ 1.73) & 54.77 ($\pm$ 12.23) & 45.04 ($\pm$ 1.07) & 37.88 ($\pm$ 14.05) & 12.95 ($\pm$1.49) &24.1 ($\pm$2.75) \\
& 3 & 55.23 ($\pm$ 4.94) & 64.76 ($\pm$ 0.99) & 45.25 ($\pm$ 1.51) & 45.08 ($\pm$ 6.17) & 17.08 ($\pm$1.32)	&32.78 ($\pm$1.79) \\
& 5 & 59.86 ($\pm$ 0.93) & 64.89 ($\pm$ 1.63) & 47.37 ($\pm$ 1.33) & 46.96 ($\pm$ 3.75) & 18.26 ($\pm$0.91) &	35.44 ($\pm$1.85) \\
\midrule
\multirow{ 1}{*}{\textbf{Flan-T5}} 
& 5 & 30.32 ($\pm$6.62) & 29.33 ($\pm$1.8) & 38.84 ($\pm$4.23) & 30.68 ($\pm$12.53) & 14.74 ($\pm$6.78) & 4.84 ($\pm$1.32) \\
\bottomrule

\end{tabular}
}
\caption{Few-shot scores with $k = \{1, 3$ $and$ $5\}$. We ran experiments with 3 seeds and averaged the results. Results show F1 scores and standard deviation. We have chosen the format that works best for each dataset. CHEM, CDR, MEDM, NCBI on \emph{text input and JSON output} and PICO and CHIA with \emph{text input and code output}, with an exception of Claude 2 which we experimented on JSON.}
\label{tab:fewshot}
\end{table*}

\subsection{Few-Shot Experimental Setup}
\label{ssec:fewshot}

For our few-shot experiments, we adopt the combination of input/output formats that performed the best for each dataset in the zero-shot setting. 
We validated this decision by evaluating all combinations of input/output formats on one of the datasets (i.e., CDR) and observing that the best performing format in zero-shot also applies to the few-shot setting (for $k=\{1, 3, 5\}$). These results are shown in Table~\ref{tab:fewformat} of the Appendix \ref{apx:fewshotoutput}.


In addition to input/output formats, few-shot prompting can also vary along two axes: (i) Selection of few-shot exemplars, and; (ii) Ordering of chosen exemplars. 
For the former, we compared selection of few-shot exemplars at random to the similarity-based approach from \cite{gutiérrez2022thinking}.
For the latter, we compared passing exemplars in a random but fixed order against shuffling exemplars per test instance. 
In preliminary experiments, we did not observe meaningful differences in performance based on these strategies. 
Therefore, we executed the rest of the experiments with randomly selected exemplars shuffled per test instance. 
See Appendix~\ref{apx:fewshotsorder} for additional details on these few-shot prompting strategies. 

We evaluate all the models for ${k=\{1, 3, 5\}}$ and report the average performance across three seeds (additional results for larger values of $k$ are provided in Figure~\ref{fig:fewshotval}).  

\subsection{Fine-tuning Experimental Setup }
\label{ssec:finetune}

To put our results into context, we also measure the performance of a smaller language model fine-tuned on the each of the datasets. Specifically, we fine-tune Flan-T5 XL on linearized targets.
We train the model on the same set of 5 instances used in the few-shot experiments using LoRA, a parameter efficient fine-tuning method~\cite{hu2021lora}. We provide implementation details in Appendix ~\ref{apx:implementation}.



\subsection{Results}
\label{sec:results}

In preliminary experiments, 
we see that prompts augmented with schema definitions perform worse across all models and datasets. As for output formats, we find that JSON was preferred on most datasets with the exception of PICO and CHIA. This observation holds consistently across all models. 
See Table \ref{tab:zeroshot} for the results of GPT-3.5, Claude 2 and Llama 2 on all datasets. 

These findings motivate our few-shot setup, in which we unsurprisingly find that performance tends to increase with the number of shots (Table \ref{tab:fewshot}).
Finally, we see that few-shot learning with instruction tuned LLMs dramatically outperforms a small LM fine-tuned on the same 5 instances.

\begin{table*}[h]
\centering
\small
\begin{tabular}{llcccccc} 
\toprule
\textbf{Model} & \textbf{Setting} & \textbf{CHEM} & \textbf{CDR} & \textbf{MedM} & \textbf{NCBI} & \textbf{PICO} & \textbf{CHIA}\\
\midrule
\multirow{4}{*}{\textbf{GPT-3.5}} 
& \textbf{ZS} & 48.61 & 67.65&	43.77&	54.05 &	10.25	& 7.50 \\
& \textbf{+Def} & \cellcolor{red!25}48.34	(-0.27) & \cellcolor{green!25}68.21 (+0.56)	& \cellcolor{green!25}45.00 (+1.23)	& \cellcolor{red!25}51.94	(-2.11) & \cellcolor{red!25}10.20 (-0.05)	& \cellcolor{green!25}7.95 (+0.45) \\
& \textbf{IP} & \cellcolor{red!25}47.27 (-1.34) & \cellcolor{red!25}66.12 (-1.53) & \cellcolor{red!25}42.71 (-1.06) & \cellcolor{red!25}51.18 (-2.87) & \cellcolor{green!25}10.27 (+0.02) & \cellcolor{green!25}7.59 (+0.09) \\
& \textbf{+Def} & \cellcolor{green!25}56.39 (+7.78)	& \cellcolor{green!25}72.86 (+5.21) & \cellcolor{green!25}50.05 (+6.28)	& \cellcolor{green!25}58.24 (+4.19) & \cellcolor{red!25}9.88 (-0.37) & \cellcolor{green!25}17.64 (+10.14) \\ \midrule
\multirow{4}{*}{\textbf{Claude 2}} 
& \textbf{ZS} & 54.28 &	70.07 &	36.98 &	44.17 &	7.26 & 20.12 \\
& \textbf{+Def} & \cellcolor{green!25}57.62 (+3.34) & \cellcolor{red!25}68.91 (-1.16)	& \cellcolor{red!25}36.12 (-0.86) & \cellcolor{red!25}43.65 (-0.52) & \cellcolor{green!25}7.67 (+0.41) & \cellcolor{red!25}19.17 (-0.95) \\
& \textbf{IP} & \cellcolor{red!25}52.93 (-1.35) & \cellcolor{red!25}69.34 (-0.73) & \cellcolor{red!25}36.71 (-0.27) &	\cellcolor{red!25}43.43 (-0.74) & \cellcolor{green!25}7.66 (+0.40) & \cellcolor{red!25}19.82 (-0.30) \\
& \textbf{+Def} & \cellcolor{green!25}59.96 (+5.68) & \cellcolor{green!25}73.04 (+2.97) & \cellcolor{green!25}41.82 (+4.84) & \cellcolor{green!25}51.60 (+7.43) & \cellcolor{green!25}8.98 (+1.72) & \cellcolor{green!25}22.12 (+2.00) \\ \midrule
\multirow{4}{*}{\textbf{Llama 2}} 
& \textbf{ZS} & 60.30 & 64.07 & 25.98 &	47.38 &	7.88 & 4.24 \\
& \textbf{+Def} & \cellcolor{green!25}67.49 (+7.19) & \cellcolor{green!25}68.54 (+4.47) & \cellcolor{green!25}35.56 (+9.58) & \cellcolor{green!25}51.44 (+4.06) & \cellcolor{green!25}8.54 (+0.66) & \cellcolor{green!25}9.50 (+5.26) \\
& \textbf{IP} & \cellcolor{red!25}58.31 (-1.99) & \cellcolor{red!25}65.63 (-1.56) & \cellcolor{red!25}24.54 (-1.44) & \cellcolor{red!25}45.58 (-1.80) & \cellcolor{red!25}7.49 (-0.39) & \cellcolor{green!25}4.50 (+0.26) \\
& \textbf{+Def} & \cellcolor{green!25}67.54 (+7.24) & \cellcolor{green!25}69.05 (+4.98) & \cellcolor{green!25}34.90 (+8.92) & \cellcolor{green!25}50.57 (+3.19) & \cellcolor{green!25}9.59 (+1.71) & \cellcolor{green!25}9.42 (+5.18) \\ \midrule
\multirow{4}{*}{\textbf{GPT-4}} 
& \textbf{ZS} & 62.12 & 70.92 & 47.13 & 54.67 & 7.29 & 16.39 \\
& \textbf{+Def} & \cellcolor{green!25}67.05 (+4.93)	& \cellcolor{green!25}76.19 (+5.27) & \cellcolor{green!25}51.91 (+4.78) & \cellcolor{green!25}60.91 (+6.24) & \cellcolor{green!25}9.24 (+1.95) & \cellcolor{green!25}20.88 (+4.49) \\
& \textbf{IP} & \cellcolor{red!25}59.67 (-2.45) & \cellcolor{red!25}69.41 (-1.51) & \cellcolor{red!25}47.01 (-0.12) & \cellcolor{red!25}52.31 (-2.36) & \cellcolor{green!25}7.47 (+0.18) & \cellcolor{green!25}17.94 (+1.55) \\
& \textbf{+Def} & \cellcolor{green!25}65.39 (+3.27) & \cellcolor{green!25}75.62 (+4.70) & \cellcolor{green!25}52.13 (+5.00) & \cellcolor{green!25}58.72 (+4.05) & \cellcolor{green!25}9.47 (+2.18) & \cellcolor{green!25}20.09 (+3.70) \\
\bottomrule
\end{tabular}
\caption{Zero-shot (ZS) scores with Definition Augmentation (+Def), Iterative Prompting (IP) and Iterative Prompting augmented with Definitions (+Def) on four models. Results show F1 scores and the delta wrt zero-shot in the parenthesis.}
\label{tab:defzeroshot}
\end{table*}

\begin{table*}[ht]
\centering
\small
\resizebox{!}{2cm}{
\begin{tabular}{llcccccc} 
\toprule
\textbf{Model} & \textbf{Setting} & \textbf{CHEM} & \textbf{CDR} & \textbf{MedM} & \textbf{NCBI} & \textbf{PICO} & \textbf{CHIA}\\
\midrule
\multirow{2}{*}{\textbf{GPT-3.5}}  & 
\textbf{FS} & 57.92 ($\pm$ 0.78) &	68.89 ($\pm$ 1.03) &49.08 ($\pm$ 01.33) &56.02 ($\pm$ 1.48)	&  11.07 ($\pm$ 1.77)	& 21.72 ($\pm$ 1.23) \\
& \textbf{+Def} & \cellcolor{green!25}59.23 ($\pm$ 1.54) & \cellcolor{red!25}68.7 ($\pm$ 2.47) & \cellcolor{red!25}48.41 ($\pm$ 0.77) & \cellcolor{green!25}57.6 ($\pm$ 2.75) & \cellcolor{green!25}11.19 ($\pm$ 0.52) & \cellcolor{green!25}22.15 ($\pm$ 1.03)\\

\hline
\multirow{ 2}{*}{\textbf{Claude 2}} 
 & \textbf{FS} & 61.6 ($\pm$ 0.36)	& 71.95 ($\pm$ 2.62)	& 48.3 ($\pm$ 1.44)	& 44.92 ($\pm$ 1.62)	& 6.2 ($\pm$ 2.83)	& 19.72 ($\pm$ 2.94) \\
   & \textbf{+Def} & \cellcolor{red!25}61.17 ($\pm$ 0.26) & \cellcolor{green!25}72.81 ($\pm$ 1.58)	& \cellcolor{green!25}49.32 ($\pm$ 1.36) & \cellcolor{green!25}48.98 ($\pm$ 1.51)	& \cellcolor{green!25}9.97 ($\pm$ 2.13) & \cellcolor{green!25}22.21 ($\pm$ 1.03)\\

\hline
\multirow{ 2}{*}{\textbf{Llama 2}} 
 & \textbf{FS} & 60.15 ($\pm$ 0.92)	& 66.77 ($\pm$ 1.32)	& 38.92 ($\pm$ 11.83) & 47.97 ($\pm$ 3.65) & 8.0 ($\pm$1.98) 	&	9.32 ($\pm$ 0.45) \\
   & \textbf{+Def} & \cellcolor{red!25}59.86 ($\pm$ 0.93) & \cellcolor{red!25}64.89 ($\pm$ 1.63)	& \cellcolor{green!25}47.37 ($\pm$ 1.33) & \cellcolor{red!25}46.96 ($\pm$ 3.75)	& \cellcolor{green!25}18.26 ($\pm$ 0.91) & \cellcolor{green!25}35.44 ($\pm$ 1.85)\\
 \hline
\multirow{ 2}{*}{\textbf{GPT-4}} 
 & \textbf{FS} & 64.92 ($\pm$ 1.28)	& 74.23 ($\pm$ 3.48)	& 54.59 ($\pm$ 1.89)	& 62.28 ($\pm$ 1.97)	& 8.74 ($\pm$ 1.68)	&  23.21 ($\pm$ 1.60) \\
   & \textbf{+Def} & \cellcolor{green!25}69.72 ($\pm$ 0.68)& \cellcolor{green!25}79.63 ($\pm$ 2.96) & \cellcolor{green!25}59.17 ($\pm$ 1.5)	& \cellcolor{green!25}66.21 ($\pm$ 0.96)& \cellcolor{red!25}7.63 ($\pm$ 0.58) & \cellcolor{green!25}24.51 ($\pm$ 0.77)\\

\hline
\end{tabular}
}
\caption{Few-shot scores with Definition Augmentation (+Def) with $k$ = 5. We ran experiments with 3 seeds and averaged the results. Results show F1 scores and standard deviation in the parenthesis.}
\label{tab:deffewshot}
\end{table*}

\section{Augmenting Prompts with Definitions} 
\label{sec:defaug}
ICL approaches rely on the parametric knowledge acquired by the models during pre-training. However, this internal knowledge can be incorrect, insufficient, or outdated.
Prior work has tried to address knowledge gaps in LLMs by augmenting prompts with relevant factual knowledge \emph{on-the-fly}, improving performance on language understanding tasks like question answering~\cite{baek2023knowledge,wang2023augmenting}.

This motivates us to explore 
whether dynamically augmenting prompts with relevant knowledge improves ICL performance for biomedical NER. 
In our work, we focus on a specific category of knowledge---\emph{definitions of biomedical concepts} present in the input text. 
Intuitively, generic LLMs may not be proficient with  biomedical concepts; providing targeted information at test time may permit fast adaptation to this domain. 

We propose to operationalize this approach as follows. 
First, we curate a knowledge base of biomedical concept definitions and leverage an off-the-shelf entity linker to map occurrences of concepts to entries in the knowledge base (\S \ref{ssec:defaugexpsetup}). Second, we perform inference with a sequence of prompts: We prompt models to extract entities as discussed in \S \ref{sec:ICLforNER}, and then craft follow-up prompts augmented with concept definitions that ask the model to revise initial extractions. 
Revisions can remove or add entities, or re-assign entity types. 
We provide definitions for all the entities identified by the model in the first turn, and all other biomedical concepts that can be linked to the knowledge base (as identified by the entity linker). 

We hypothesize that adding definitions for LLM-extracted entities may improve precision (original model extractions could be corrected) and adding definitions for other noun phrases can improve recall (model recognizes potential entities that were missed in the first pass).\footnote{ 
We also tested the pipeline without adding definitions for other noun phrases (i.e., removing potential recall improvements) and observed smaller improvements in performance compared to our overall approach.} We evaluate this approach in zero-shot (\S \ref{ssec:zsdefaug}) and few-shot (\S \ref{ssec:fsdefaug}) settings.

\subsection{Concept Definitions}
\label{ssec:defaugexpsetup}

We obtain concept definitions from Unified Medical Language System (UMLS), a collection of key terminology and coding standards from several biomedical vocabularies, standards and knowledge bases~\cite{bodenreider2004unified}. 

Some concepts in UMLS belong to fairly broad categories (e.g., event, activity, group) and their definitions might not provide much utility to LLMs. We avoid including definitions for such concepts by curating a set of fine-grained categories where two of the authors independently went through the entire list of 127 semantic types in UMLS and discarded generic ones (e.g., 'Plant', 'Chemical') which did not require additional biomedical knowledge to comprehend. All types retained by both authors were included in the final list, provided in the Appendix \ref{tab:tuis}. Note that some entities do not have definitions in either UMLS nor Wikipedia. For such entities (about 10\% of all entities in each dataset), we do not provide any definitions. 

At inference time, we use the entity linker available in the SciSpaCy package~\cite{neumann-etal-2019-scispacy} to map all mentions of biomedical concepts in the input text to entries in UMLS, and retrieve the associated definitions.



\subsection{Zero-Shot Definition Augmentation}
\label{ssec:zsdefaug}


In the zero-shot setting, we first prompt the model to extract entities as described in \S \ref{ssec:zeroshot}. Then we consider two strategies for follow-up prompting.
\paragraph{Single-turn (ZS+Def):} A single definition augmented follow-up prompt asks the model to make corrections to all extracted entities. 
\paragraph{Iterative Prompting (IP+Def):} Iterative prompts augmented with the definition of a single concept and asking the model to make corrections to a single extracted entity (if needed) at a time.
This breaks down the correction process into atomic steps, but significantly increases the number of inference steps (which incurs additional costs when using proprietary models). 

Our approach is related to prior work suggesting that LLMs are able to correct and revise their own outputs and this self-verification can improve performance in clinical information extraction tasks {\cite{Gero2023SelfVerificationIF}}. The novelty on offer here is providing contextual knowledge to aid the process of self-verification. In our experiments, we ablate the impact of self-verification from that of the concept definitions.


\subsection{Few-Shot Definition Augmentation}
\label{ssec:fsdefaug}

In the few-shot setting, again we first prompt the model to extract entities as described in \S \ref{ssec:fewshot}, and then ask it to correct the extractions in a follow-up prompt with concept definitions. The follow-up prompt includes: (i) all few-shot exemplars provided in the first prompt along with the associated concept definitions; and (ii) definitions for all the concepts identified in the current input (both for extracted entities and other biomedical concepts). 

Here, we only test the single-turn strategy because including few-shot examples dramatically increases context size, rendering iterative prompting prohibitively expensive. 


\begin{figure*}[!ht]
\centering
\begin{subfigure}[b]{0.48\textwidth}
    \centering
    \includegraphics[width=\textwidth]{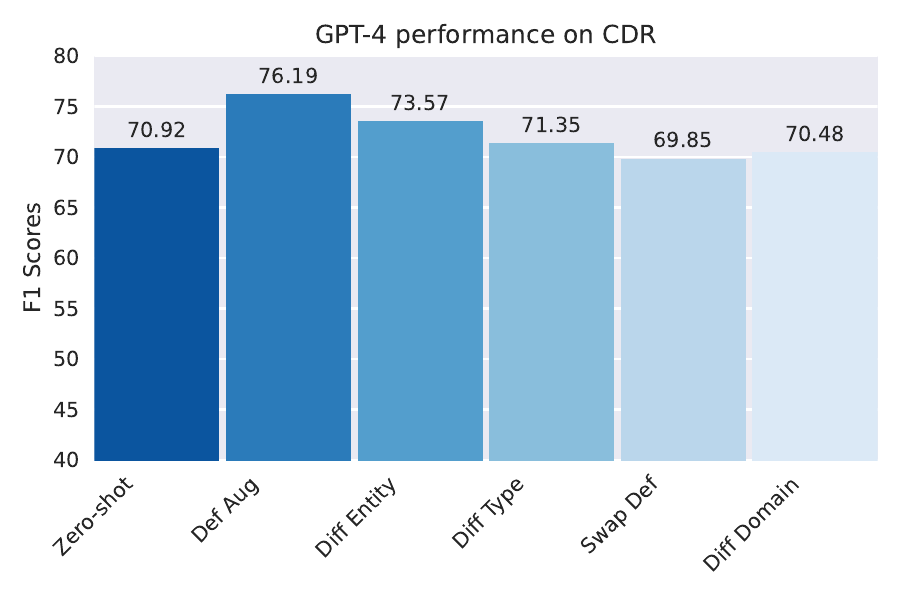} 
\end{subfigure}
\hfill
\begin{subfigure}[b]{0.48\textwidth}  
    \centering 
    \includegraphics[width=\textwidth]{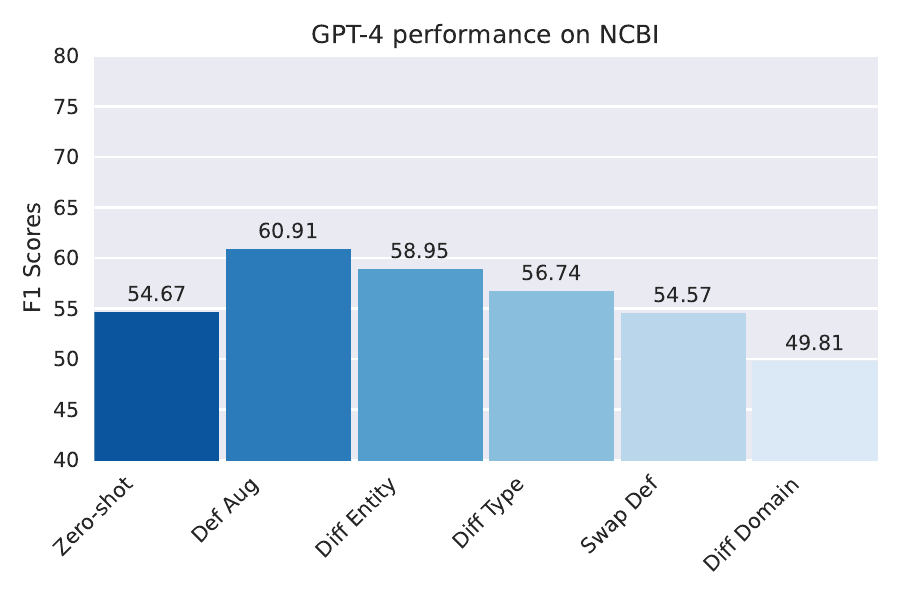}   
\end{subfigure}
\vskip\baselineskip
\begin{subfigure}[b]{0.48\textwidth}
    \centering
    \includegraphics[width=\textwidth]{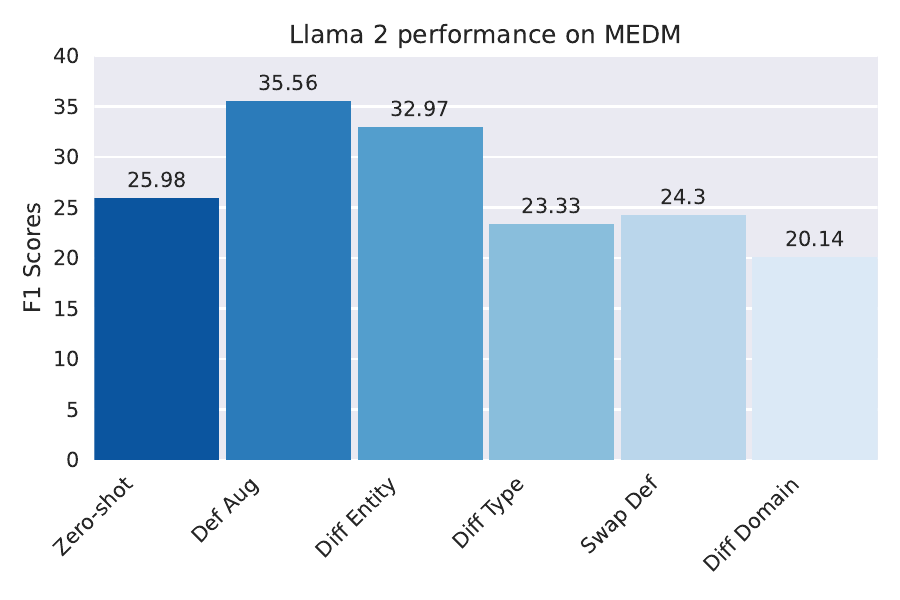}  
\end{subfigure}
\hfill
\begin{subfigure}[b]{0.48\textwidth}  
    \centering 
    \includegraphics[width=\textwidth]{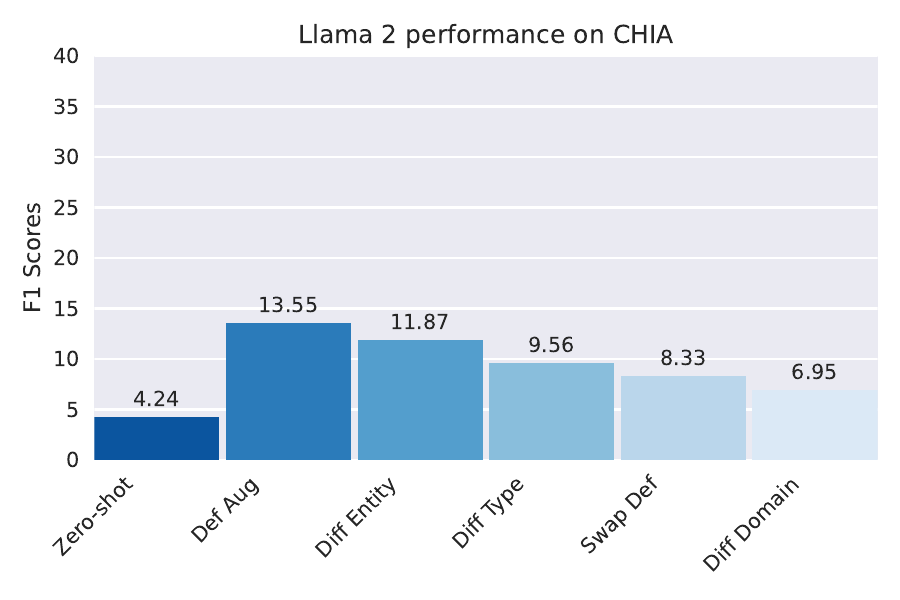}   
\end{subfigure}

\caption {Definition relevance ablations with GPT-4 on CDR dataset (top-left), NCBI (top-right) and Llama 2 on MEDM dataset (bottom-left) and CHIA (bottom-right). We see similar trends across all models and datasets - a consistent decrease in performance with less relevant definitions.}

\label{fig:defplots}
\end{figure*}

\subsection{Definition Augmentation Results}
\label{ssec:resultsdefaug}

All experiments are carried out with JSON outputs to maintain a uniform experimental setting across all datasets. 
The few-shot experiments are executed with $k=5$ shots randomly selected and shuffled per test instance. 
We run each experiment with three different random seeds and report average performance. 
In addition to the models considered in the previous section, here we also evaluate GPT-4---this is motivated by prior work suggesting that GPT-4 is more competent that GPT-3.5 at editing previous outputs~\cite{Gero2023SelfVerificationIF}, which is a key step in our proposed approach. 
However, given the high costs of querying the API, we subsampled our test sets to 100 instances for these experiments. Tables~\ref{tab:defzeroshot} and~\ref{tab:deffewshot} present the performance of GPT-3.5, Claude 2, Llama 2 and GPT-4 with definition augmentation on all datasets in the zero- and few-shot settings, respectively. 

In zero-shot settings, we see consistent and significant improvements in the performance of Llama 2 and GPT-4 with both prompting strategies. We observe an average increase of  32.6\% and 33.9\% for Llama 2 and 15\% and 13.7\% for GPT-4 using single turn and iterative prompting, respectively. 
However, Claude 2 and GPT-3.5 
only benefit when using the iterative prompting approach, with average gains of 12\% and 29.5\%, respectively. 
We also assessed the performance of iterative prompting but \emph{without} the definitions---this is similar to the \citet{Gero2023SelfVerificationIF} self-verification method. 
However, our results show that the models are not able to correct their predictions in the absence of  definitions.

In the few-shot setting, we also see improvements in most cases. 
Claude 2 and GPT-4 improve on 5 of 6 datasets; Llama 2 and GPT-3.5 show gains on 3 and 4 datasets, respectively. 
Overall, we found that GPT-4 with iterative prompting achieves the best performance.

Our results show that concept definition augmented prompts improve the performance of biomedical NER. 
A key step of this approach is linking biomedical concepts to definitions in UMLS. 
One natural question is how much of the observed gains are simply due to the use of an entity linking model which was explicitly trained to recognize entities.
To answer this, we first measured the performance of the entity linker by itself on the same test sets and found that it performs poorly, with an average F1 of 1.05 across all the datasets. Then, to verify that LLM is not just copying candidate entities identified by the entity linker, we conducted an ablation where we simply add the candidate entities \emph{without} the corresponding concept definitions. The results in Table \ref{tab:gptuniablation} show that this is not as effective as our proposed approach and in some cases underperforms compared to the zero-shot baseline.

\section{Assessing the Utility of Definition Knowledge}
\label{sec:defablations}

We further assess the utility of concept definitions by conducting ablation experiments probing the following dimensions: (1) Relevance of concept definitions; 
and (2) Source of definition knowledge. 

We conduct all experiments in the single-turn zero-shot setting (\S \ref{ssec:zsdefaug}), with one closed model (GPT-4) and one open-source model (Llama 2), on the two datasets with the largest gains in performance from concept definitions (CDR and NCBI for GPT-4; MEDM and CHIA for Llama 2).



\subsection{Probing Definition Relevance}
\label{ssec:defrelevance}

Motivated by prior work showing that LLMs often produce correct predictions even with misleading or irrelevant prompts \cite{webson-pavlick-2022-prompt}, we ablate over the \emph{relevance} of definitions provided for a given entity. 
This 
allows us to assess whether performance gains are due to accurate definitions or simply from additional context, irrespective of relevance. 
To this end, we measure the performance of increasingly \emph{less} relevant knowledge by swapping out various components of provided definitions. 
These ablations are realized as follows.  



\vspace{0.2em}
\noindent
\textbf{Diff Entity} include definitions of concepts mentioned in a different instance (within the same dataset). As this samples instances in the same dataset, it will include concepts from the same entity types being extracted (e.g., for NCBI, the swapped concepts will include some diseases). 

\vspace{0.2em}
\noindent
\textbf{Diff Type} include definitions from concepts mentioned in a different instance within the same dataset, but exclude concepts from the entity types being extracted (e.g., for NCBI, add all swapped concepts that are not diseases). 

\vspace{0.2em}
\noindent \textbf{Swap Def} replace definitions for all concepts mentioned in the current instance with random incorrect definitions (e.g., for NCBI, if the disease extracted is Arrhythmia, we provide an incorrect definition for Arrhythmia).

\vspace{0.2em}
\noindent \textbf{Diff Domain} include definitions for concepts mentioned in an instance from a \emph{different domain}. For instance, for datasets containing Pubmed abstracts (MEDM), we add concepts mentioned in a dataset of clinical trial criteria (CHIA) and vice versa.

Figure~\ref{fig:defplots} shows the performance of GPT-4 and Llama 2 under different definition relevance ablations on these datasets.
We see similar trends across all models and datasets: A consistent decrease in performance with less relevant definitions. This provides evidence that the model is indeed capitalizing on the definitions and suggests that the quality of the definitions plays a critical role on our proposed method. Interestingly, we observe that augmenting prompts with definitions of other entities (of the same type) also yields consistent gains across models and datasets. 
We are unsure what explains this, though perhaps because the entities are of the same type, they are similar enough for the model to make use of the definitions. 
Finally, we do observe some gains from definitions of entities of a different type, but these are smaller and less consistent.

%



\begin{table}
\centering
\small
\begin{tabular}{ lcccc } 
\toprule
 \textbf{Setting} & \textbf{CDR} & \textbf{NCBI} & \textbf{MEDM} & \textbf{CHIA} \\
\midrule
 \textbf{ZS} & 70.92 & 54.67 & 25.98 & 4.24 \\ 
 \textbf{Def Aug} & 76.19 & 60.91 & 35.56 & 9.50 \\ 
 \textbf{Only ents} & 68.14 & 47.29 & 28.92 & 7.48\\ 
 \hline
\end{tabular}
\caption{Ablations with GPT-4 [CDR, NCBI] and Llama 2 [MEDM, CHIA], providing only the entities without the definitions. }
\label{tab:gptuniablation}
\end{table}

\subsection{Probing Definition Sources}
\label{ssec:defsource}

After establishing that the success of our approach is largely due to adding relevant definition knowledge, we assess the impact of the \emph{source} of definition knowledge. We evaluate the same models and datasets as in the previous experiments but using concept definitions: (i) collected from Wikidata; and (ii) automatically generated by GPT-4.


Table~\ref{tab:gptsourcedef} shows the results for all models and data sources.
We observe that definitions from Wikidata also improve over the zero-shot baseline, albeit to a lesser degree than UMLS. On the other hand, the definitions generated by GPT-4 seem to have little to no impact on the model's performance. These results again highlight the importance of the knowledge source: we see larger improvements with concept definitions from a more domain-specific source. However, seeing that models can also benefit from concept definitions from more general sources such as Wikidata, suggests that our proposed approach may also be suitable for applications in other, less specialized, domains.





\begin{table}
\centering
\small
\begin{tabular}{ lcccc } 
\toprule
 \textbf{Setting} & \textbf{CDR} & \textbf{NCBI} & \textbf{MEDM} & \textbf{CHIA} \\
\midrule
 \textbf{ZS} & 70.92 & 54.67 & 25.98 & 4.24 \\ 
 \textbf{+UMLS} & 76.19 & 60.91 & 35.56 & 9.50 \\ 
 \textbf{+Wiki} & 72.9 & 57.5 & 32.6 & 9.53\\ 
 \textbf{+GPT-4} & 69.24 & 54.83 & 25.29 & 7.32 \\ 
 \hline
\end{tabular}
\caption{Ablations with GPT-4 [CDR, NCBI] and Llama 2 [MEDM, CHIA], providing definitions from different sources. Original source being \textbf{UMLS} and ablations with Wikipedia and GPT-4 generated definitions. }
\label{tab:gptsourcedef}
\end{table}

\section{Related Work}
\label{sec:background}

\par{\textbf{Information Extraction with LLMs} Recent work has shown that LLMs are capable of extracting information from documents in zero- and few-shot settings. 
For instance, \citet{agrawal2022large} found that GPT-3 competes with or outperforms smaller models on a small set of clinical tasks extraction tasks. However, in the scientific and biomedical domain, LLMs 
underperformed relative to fine-tuned models \cite{gutiérrez2022thinking}. 
GPT-3's ICL \cite{brown2020language} compares favorably to supervised models on many standard NLP tasks (e.g., NLI, text classification, machine translation \cite{liu-etal-2022-makes}). 
Several methods have been introduced to improve its performance, optimizing prompt retrieval \cite{shin-etal-2021-constrained}, ordering \cite{lu2022fantastically}, and design \cite{perez2021true}}.

\par\textbf{Iterative Prompting with LLMs} 
In recent work, \citet{Gero2023SelfVerificationIF}, used self-verification to improve clinical information extraction by iteratively prompting a LLM to sequentially identify entities, detect missing entities, ground the extractions in evidence (i.e., specific spans in the input), and remove incorrect extractions. 

This builds on prior works that iteratively prompt LLMs to improve their performance~\cite{wu2022ai,wang-etal-2022-iteratively}.

\par\textbf{Knowledge Augmentation with LLMs}
Prior to LLMs, REALM \cite{guu2020realm} and RAG \cite{lewis2021retrievalaugmented} proposed to integrate knowledge by  retrieving documents from unstructured corpora (e.g., Wikipedia) and facts from Knowledge Graphs (KGs), and conditioning outputs on these. 

Recently, concurrent to this work, \citet{nori2023generalist} explores iterative prompting with knowledge augmentation in clinical domain. Their prompting strategy combines kNN-based few-shot example selection, GPT-4–generated chain-of-thought prompting, and answer-choice shuffled ensembling to reduce the error of rate medical question answering (MedQA) by 27\%. 
\section{Conclusions}
\label{sec:conclusion}

In this work, we extensively evaluated the performance of ICL approaches for biomedical NER with modern LLMs. We compared different combinations of input and output formats and characterized the main types of errors made by these models. 
We then proposed and evaluated a method for rapid adaptation of general LLMs to biomedical NER tasks by providing models with \emph{concept definitions} from an external knowledge base dynamically. 

We perform inference with a sequence of prompts, allowing models to revise their predictions given definitions of key concepts in the input. 
The first prompt asks the model to extract entities from the input; subsequent prompts are augmented with definitions for all biomedical concepts including the entities identified in the first prompt, and ask the model to revise its predictions. 

Our evaluation---conducted over 6 datasets---showed consistent and often substantial improvements over baselines, especially in zero-shot settings. 
Ablations confirm that the observed gains stem from the models' ability to capitalize on the concept definitions. 
In particular, we observe that without these definitions the models are unable to meaningfully improve their predictions. 

While we only considered datasets from a specialized domain (biomedicine), our ablations show that our approach can also be used with more general knowledge bases, such as Wikidata. 
This provides some evidence for the potential utility of this approach in other domains. 
We leave a thorough exploration of this for future work.

\section{Limitations}
\label{sec:limitations}

Since our work evaluates (some) LLMs that have been trained on undisclosed data sources, it is possible that the models have seen parts of our evaluation sets in either pre-training or instruction tuning. 
The underlying text corpora for all datasets in our NER evaluation testbed are sourced from easily accessible text collections (e.g., PubMed, AACT) and so it is quite likely that these have been seen by models during pre-training. 
However, this is (probably) not a major issue in the case of NER, because simply training on these sentences with a language modeling objective is unlikely to impart the signal necessary for NER. 

Consequently, our primary concern is potential exposure of \emph{label information} from these datasets during some form of entity-aware training or instruction tuning phase. To assess this, we provide models with the raw text and some entity labels and test whether they are able to correctly produce the remaining entities in the original format. We observe that all models failed at this, indicating that though we cannot make strong claims about data contamination, it is unlikely that models have successfully memorized these test sets. 

Another limitation of our work is that we only evaluate only on English biomedical NER corpora and did not test how well our approach would work for other languages, tasks, or domains. 
Additionally, we rely on the availability of expert-curated knowledge (UMLS) for biomedicine---however, such resources may not be readily available for for other tasks or domains. Even within biomedical NER, we test our approach on a limited number of datasets due to the experimental costs of testing proprietary LLMs, and it is possible that our approach may not work for other datasets. 

Finally, current metrics for IE tasks are not well-suited to generative models. 
We mitigate this by performing additional human evaluation, but this approach is not scalable. 









\section*{Acknowledgements}
This work was supported in part by National Science Foundation (NSF) award 1750978. We would like to thank Doug Downey and the rest of the Semantic Scholar team at AI2, as well as the reviewers, for their valuable feedback and comments that helped improve this work.

\bibliography{anthology,custom}
\bibliographystyle{acl_natbib}

\clearpage
\appendix
\section{Input Format}
\label{apx:input}

\paragraph{Selection of few-shot examples:} Prior work has shown that in-context learning can benefit from sophisticated strategies for selecting exemplars, e.g. based on diversity \cite{hongjin2022selective} or informativeness \cite{wu2023scattershot} of the samples. We defer a thorough exploration of these strategies to future work, and here focus on two relatively simple approaches: (i) \textbf{Random}, where $k$ examples are randomly sampled; and (ii) \textbf{Retrieval}, which follows \citet{gutiérrez2022thinking}. The training set is subsampled to 100 examples; then for every test instance, $k$ \emph{most similar} examples are retrieved from this pool. Similarity between examples is computed using SPECTER2 embeddings \cite{singh2022scirepeval}. 


\paragraph{Ordering of few-shot examples:} Prior work has also shown that models can be very sensitive to the order in which examples are provided for in-context learning (e.g., \citet{lu2022fantastically}), thus we compared two ordering criteria:
(i) \textbf{Fixed order}, chosen at random; and (ii) \textbf{Shuffled order} of examples per test instance. Note that for the retrieval-based shot selection, examples are provided in decreasing order of similarity~\cite{gutiérrez2022thinking}.

\section{Ablations}
\label{apx:ablations}

\subsection{Best output format in Few Shot}
\label{apx:fewshotoutput}

Ablation experiment testing multiple format combinations on CDR with $k$=1, 3 and 5 shots.
We use text as the input format as this was the best performing over def prompts across all models and all datasets. 

\begin{table}[h]
\centering
\small
\begin{tabular}{lcc} 
\toprule
\textbf{Setting}& \textbf{K}& \textbf{CDR} \\
\midrule
\multirow{ 3}{*}{\textbf{JSON}} & 1 & 64.35 \\ 
   & 3  & 65.98 \\
   & 5 &  66.26\\
\midrule
\multirow{ 3}{*}{\textbf{Code}} & 1 & 56.17 \\
   & 3 &  60.26 \\
   & 5 & 60.56\\
\bottomrule
\end{tabular}
\caption{Few-shot JSON input and code output ablations. Results show F1 scores. We evaluate combinations of input/output formats on CDR dataset and observe that the best performing format in zero-shot also applies to the few-shot setting.}
\label{tab:fewformat}
\end{table}

\subsection{Ordering shots in Few Shot}
\label{apx:fewshotsorder}

Ablations testing example selection and ordering strategies on CDR with $k$=1, 3 and 5 shots.

\begin{itemize}[topsep=0pt,leftmargin=*]
 \item \textbf{Random:} Fixed order of $k$ examples are randomly sampled.
\item \textbf{Retrieval:} For every test instance, $k$ \emph{most similar} examples are retrieved from this pool. Similarity between examples is computed using SPECTER V2 embeddings and examples are provided in decreasing order of similarity.
\item \textbf{Random + Shuffle:} Shuffling order of examples per test instance where $k$ examples are randomly sampled.
\end{itemize}

\begin{table}[h!]
\centering
\small
\begin{tabular}{lcc} 
\toprule
\textbf{Setting}& \textbf{K}& \textbf{CDR} \\
\midrule
\multirow{ 3}{*}{\textbf{Random}} & 1 & 68.25  \\ 
   & 3  & 70.93 \\
   & 5 & 72.02 \\
\midrule
\multirow{ 3}{*}{\textbf{Random + Shuffle}} & 1 & 68.06 \\
   & 3 & 70.29  \\
   & 5 & 71.93 \\
\midrule
\multirow{ 3}{*}{\textbf{Retrieved}} & 1 & 63.94 \\ 
   & 3 & 71.46 \\
   & 5 & 72.22 \\
\bottomrule
\end{tabular}
\caption{Few-shot shot selection ablations. Results show F1 scores. We do not observe meaningful differences in performance based on these strategies, therefore we carried few-shot experiments with randomly selected exemplars shuffled per test instance. }
\label{tab:fewablation}
\end{table}

\begin{figure*}[t!]
    \centering
    \begin{subfigure}[b]{0.5\textwidth}
        \centering
        \includegraphics[scale=0.19]{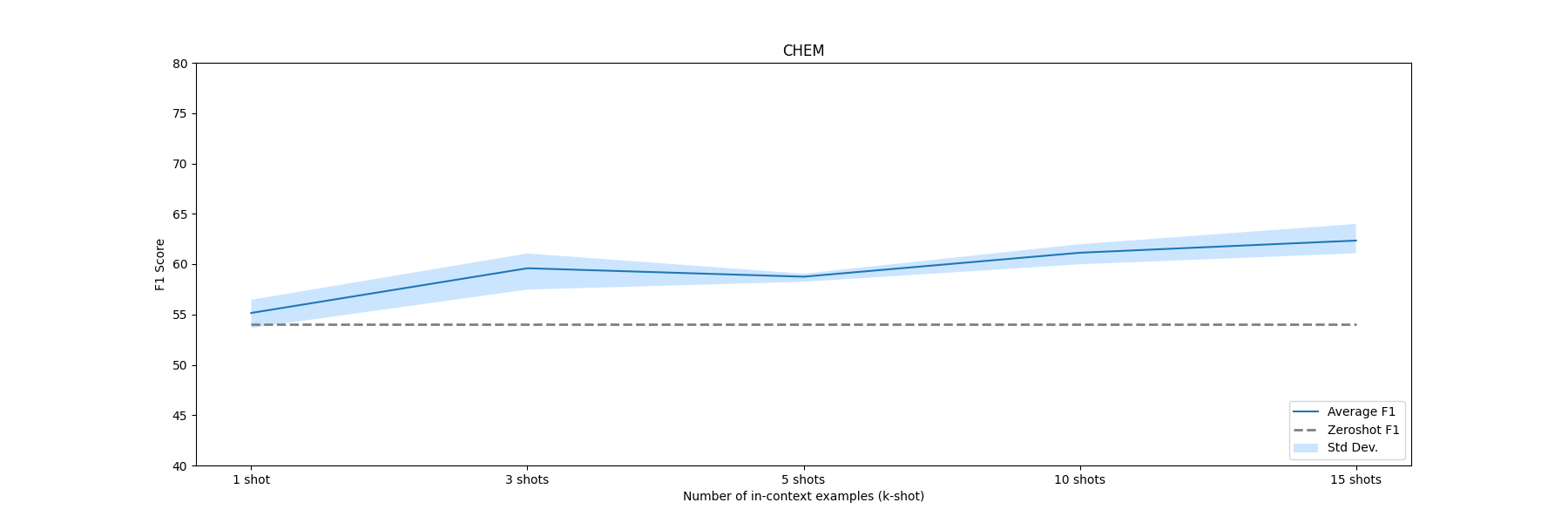}
        \caption{Few shot performance on CHEM }
    \end{subfigure}%
    \begin{subfigure}[b]{0.5\textwidth}
        \centering
        \includegraphics[scale=0.19]{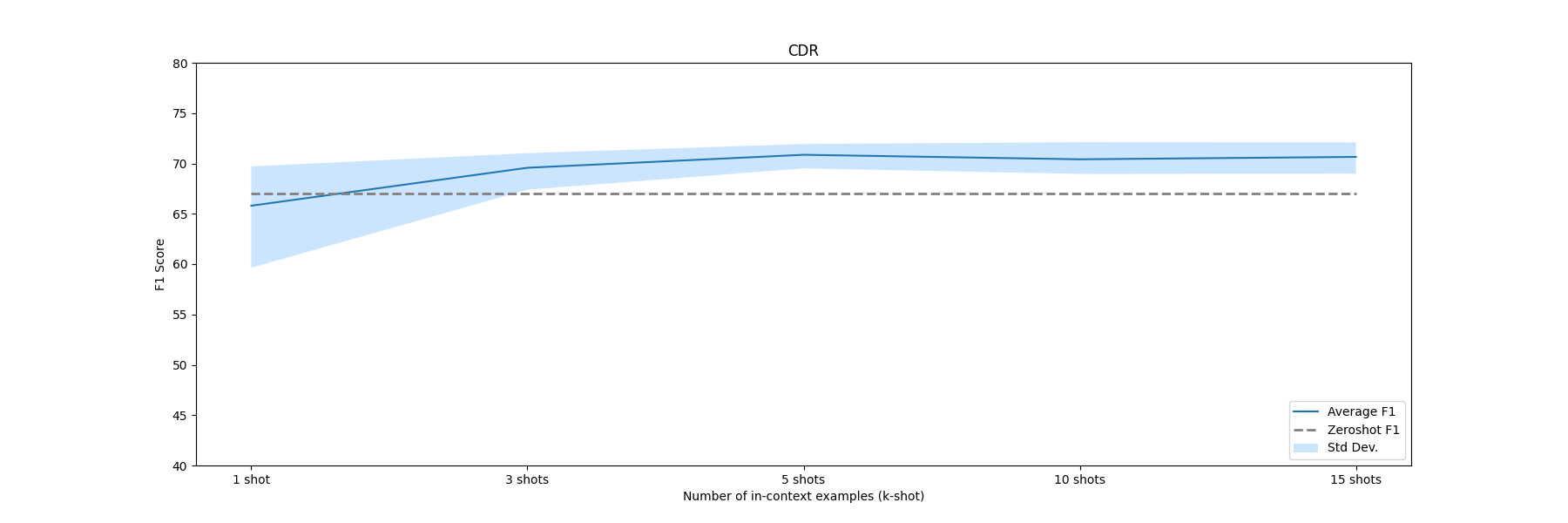}
        \caption{Few shot performance on CDR }
    \end{subfigure}
    \newline

    \begin{subfigure}[b]{0.5\textwidth}
        \centering
        \includegraphics[scale=0.19]{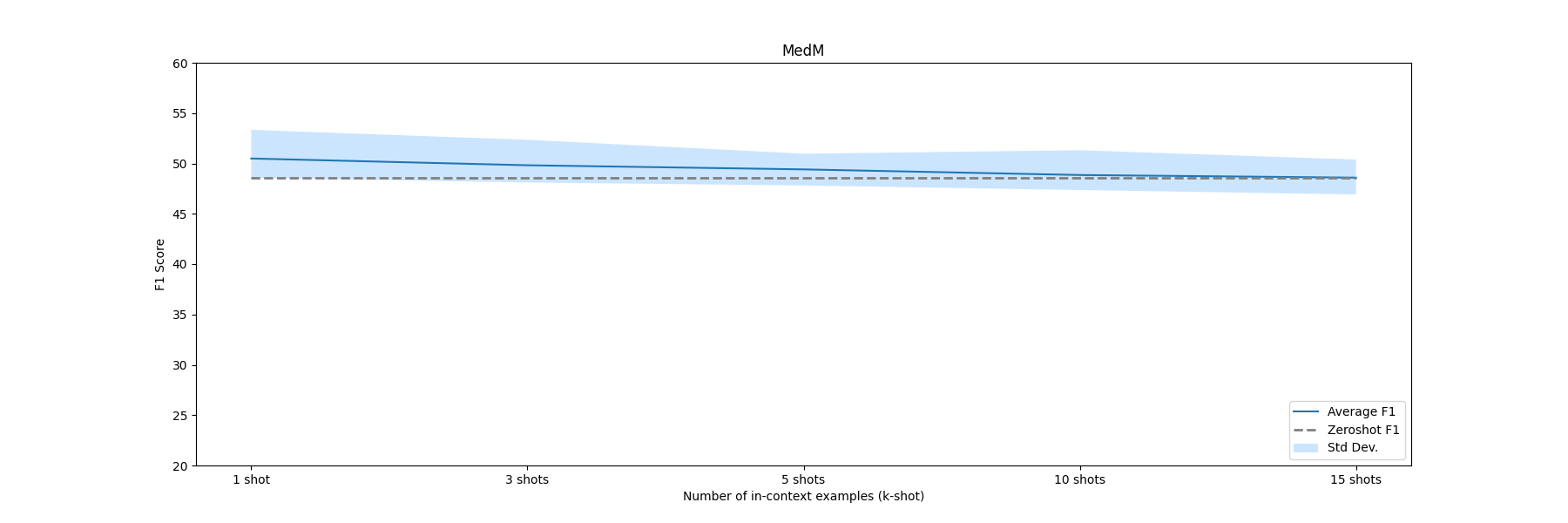}
        \caption{Few shot performance on MedM}
    \end{subfigure}%
    \begin{subfigure}[b]{0.5\textwidth}
        \centering
        \includegraphics[scale=0.19]{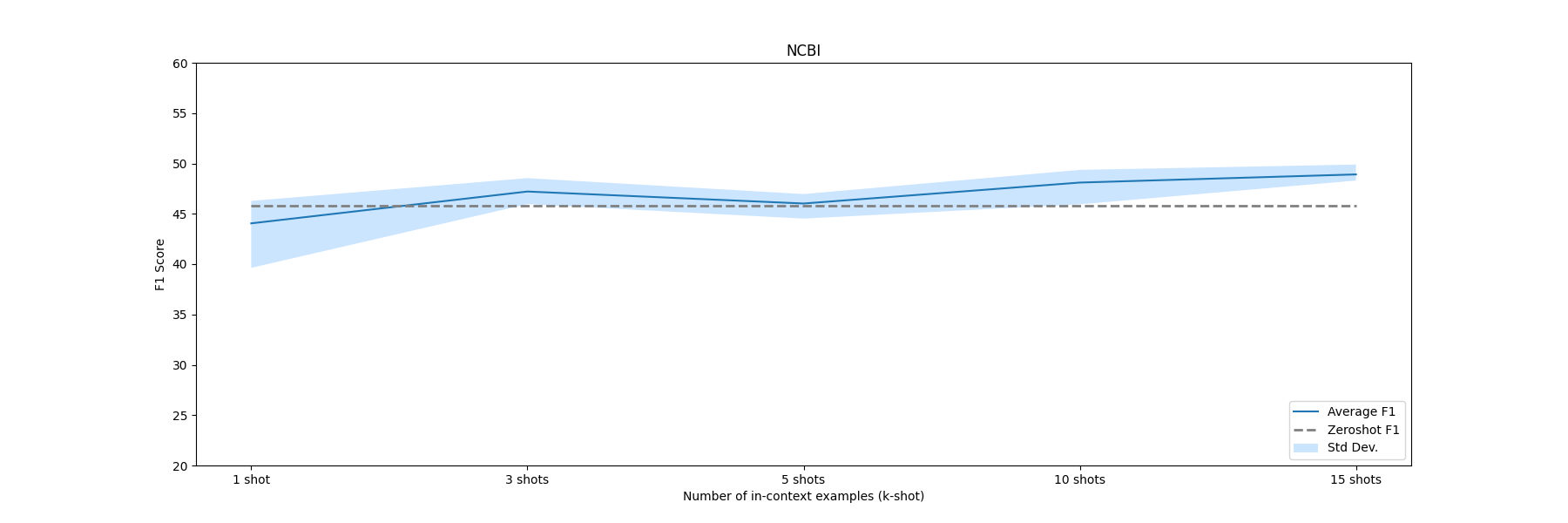}
        \caption{Few shot performance on NCBI}
    \end{subfigure}
    \newline

    \begin{subfigure}[b]{0.5\textwidth}
        \centering
        \includegraphics[scale=0.19]{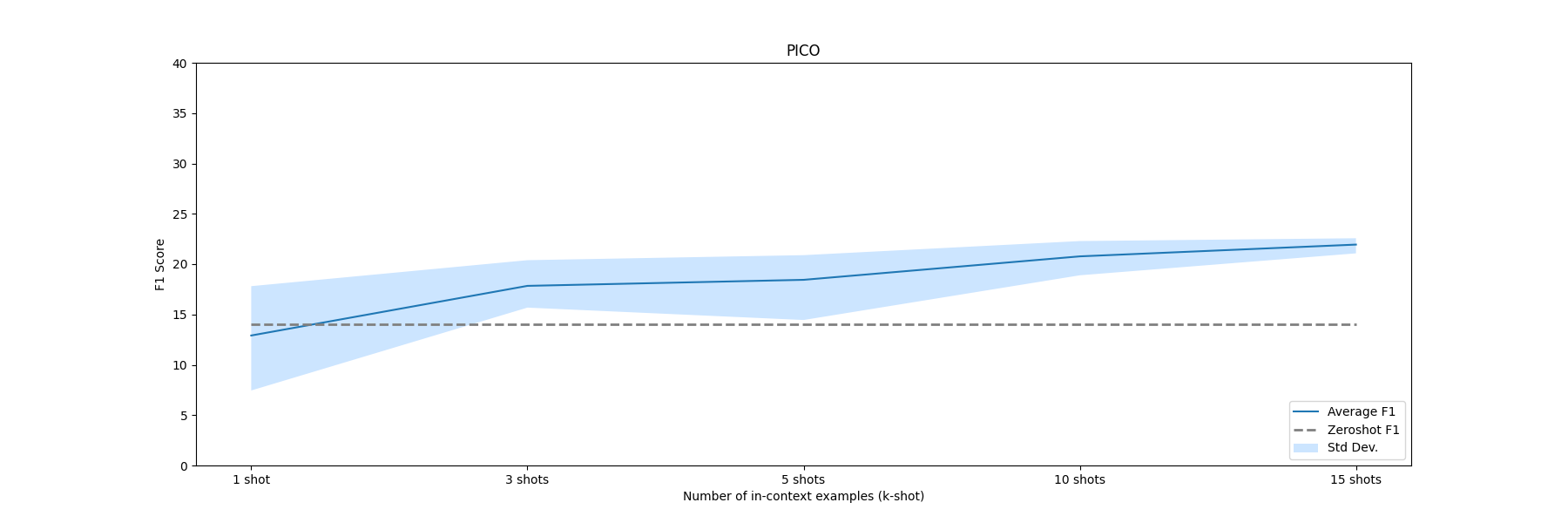}
        \caption{Few shot performance on PICO}
    \end{subfigure}%
    \begin{subfigure}[b]{0.5\textwidth}
        \centering
        \includegraphics[scale=0.19]{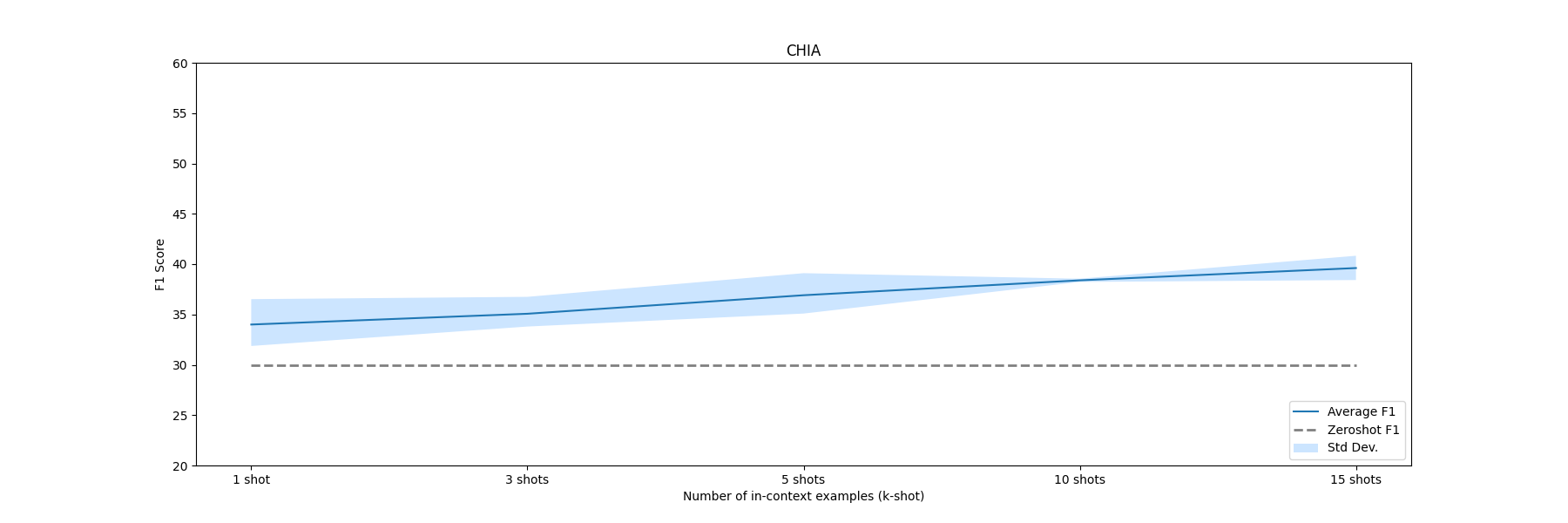}
        \caption{Few shot performance on CHIA}
    \end{subfigure}
    \caption{F1 score plotted against the number of shots in few-shot setting. Performance of all models tends to increase with the number of shots (except for NCBI and MEDM datasets where we observe minor fluctuations in performance).}
    \label{fig:fewshotval}
\end{figure*}



\section{Qualitative Error Analysis}
\label{ssec:error}

To better understand the performance of LLMs on biomedical NER and characterize errors these models still make, we conduct a qualitative error analysis of 50 examples from the best performing zero-shot and few-shot models per dataset. This analysis surfaced four major categories of errors:
\begin{itemize}[topsep=0pt,leftmargin=*]
\setlength\itemsep{-0.5em}
    \item \textbf{Type mismatch:} An entity is extracted correctly but assigned the wrong type.
    \item \textbf{Boundary issues:} The extracted entity is missing terms or contains extra terms when compared to the gold entity.
    \item \textbf{Extra entities:} Model extracts entities which are not present in gold annotations. We observe that these extractions are not always errors either, which motivates the need for human evaluation.
    \item \textbf{Missing entities:} Model does not extract entities present in gold annotation.
\end{itemize}
Table~\ref{tab:fewshoterrors} in the appendix provides an overview of the error distribution for every dataset. Several error categories mentioned above 
could potentially be corrected by providing models access to additional definition knowledge about those entities. This further motivates our exploration of definition-augmented information extraction using LLMs. 

\paragraph{Manual Evaluation}

Prior work has shown that strict F1 can underestimate the performance of generative models on information extraction tasks \cite{wadhwa2023revisiting}. To quantify the impact of this issue on our results, we conduct a small scale human evaluation on two of our datasets (i.e., PICO and CHIA) by randomly sampling 100 sentences with incorrect predictions and re-assessing all the false positive and false negatives. Our analysis showed $~51\%$ of PICO and $~30\%$ of CHIA predictions deemed incorrect were actually correct.

\begin{table*}[h]
\centering
\begin{tabular}{llcccccc} 
\toprule
\textbf{Model}  & \textbf{CDR} & \textbf{CHEM} & \textbf{MedM} & \textbf{NCBI} & \textbf{PICO} & \textbf{CHIA}\\
\midrule
\textbf{Missing Entities} & 75 & 22.6 & 47.1  & 5.5 & 10.6 & 39.2 \\

\hline
\textbf{Extra Entities}  & 14.5 & 21.3  & 14.2 & 75 & 54.54 & 11.7\\

\hline
\textbf{Boundary Issues} & 10.4 & 22.6  & 38.5 & 19.4 & 12.12 & 49\\
 \hline
\textbf{Entity Mismatch} & 0 & 33.3  & - & - & 22.7 & 0 \\

\hline
\end{tabular}
\caption{Percentage (\%) distribution of different types of errors mentioned in~\ref{ssec:error} for all datasets in zero-shot setting. Note that NCBI and MEDM datasets have only one entity type, hence there are no type mismatch errors.}
\label{tab:fewshoterrors}
\end{table*}

\begin{table*}[h]
\centering
\begin{tabular}{llcccccc} 
\toprule
\textbf{Model} & \textbf{CDR} & \textbf{CHEM} & \textbf{MedM} & \textbf{NCBI} & \textbf{PICO} & \textbf{CHIA}\\
\midrule
\textbf{Missing Entities} & 51.2 & 19.7  & 24.3 & 17  & 32.7 & 46\\

\hline
\textbf{Extra Entities} & 12.1 & 25.35  & 18.9 & 70.2  & 21.8 & 9.5\\

\hline
\textbf{Boundary Issues} & 34.1 & 28.1  & 56.7 & 12.7  & 12.7 & 44.4\\
 \hline

\textbf{Entity Mismatch} & 2.4 & 26.7 & - & - & 32.7 & 0 \\

\hline
\end{tabular}
\caption{Percentage (\%) distribution of different types of errors mentioned in~\ref{ssec:error} for all datasets in few-shot setting. Note that NCBI and MEDM datasets have only one entity type, hence there are no type mismatch errors.}
\label{tab:fewshoterrors}
\end{table*}

\section{Definition Augmentation Error Analysis}

We wanted to understand which categories of errors (as per the taxonomy in \S \ref{ssec:error}) does definition augmentation help with. For each dataset, we randomly sampled 50 instances with one or more incorrect extractions which were corrected with definition augmentation  (in the zero-shot setting). We then looked at the distribution of error types, and found that \emph{extra entities} and \emph{missing entities} were the most common error types fixed using definition information (Table \ref{tab:qualcorrectioneval}).

\begin{table}[!h]
\centering
\small
\begin{tabular}{ lcccc } 
\toprule
 \textbf{Setting} & \textbf{CDR} & \textbf{NCBI} &   \textbf{MEDM}  & \textbf{CHIA}  \\
\midrule
 \textbf{Type Mismatch} & 7.5 &  - & -  & 28.9\\ 
 \textbf{Boundary Issue} & 9.4 & 5.8  & 0 & 24 \\ 
 \textbf{Extra Entities} & 71.6 & 82.3 & 16.4 & 42\\
 \textbf{Missing Entities} & 11.3 &  11.7  & 83.5 & 4.8\\
 \bottomrule
\end{tabular}
\caption{Percentage (\%) distribution of different types of errors mentioned in~\ref{ssec:error} for 4 datasets. Note that NCBI and MEDM datasets have only one entity type, hence there are no type mismatch errors. }
\label{tab:qualcorrectioneval}
\end{table}

\begin{table}
\centering
\small
\begin{tabular}{lll}
\toprule
\textbf{Model} & \textbf{Engine}  & \textbf{Cutoff} \\
\midrule

    \textbf{GPT 3.5} & \texttt{gpt-3.5-turbo-0613}  & Sep 2021 \\

    \textbf{GPT 4} & \texttt{gpt4-0613}  & Sep 2021  \\

    \textbf{Claude 2} & \texttt{claude-2}   & Dec 2022  \\
    
    \textbf{LLaMa 2} & \texttt{llama-2-70b-chat}  & Jul 2023 \\
    
\bottomrule
\end{tabular}
\caption{Overview of all models.}
\label{tab:models}
\end{table}

\begin{table*}[t]
\centering
\small
\begin{tabular}{p{1.75cm}p{5cm}p{6cm}l}
\toprule
\textbf{Dataset} & \textbf{Descriptions}  & \textbf{Examples} \\
\midrule
    \textbf{CHEM} & The BioCreative VI Chemical-Protein Interaction corpus \cite{krallinger2017overview} contains biomedical abstracts with annotations for chemical and protein entities. &  \textbf{Sentence} : AMPK activity was measusalmon as the amount of radiolabelled phosphate transfersalmon to the SAMS peptide. \textbf{Entities} : 'Chemicals': ['phosphate'], 'Proteins': ['AMPK'] \\
    \\*

    \textbf{CDR}  & The BioCreative V Chemical-Disease Relation corpus \cite{DBLP:journals/biodb/LiSJSWLDMWL16} contains biomedical abstracts with annotations for \emph{diseases} and \emph{chemical entities}. & \textbf{Sentence} : Pre-treatment of bupivacaine-induced cardiovascular depression using different lipid formulations of propofol. 
    \textbf{Entities} : Chemicals : ['bupivacaine', 'propofol'], "Diseases": ['cardiovascular depression'] \\ 
    \\*
    
    \textbf{NCBI}  & The Natural Center for Biotechnology Information Disease corpus \cite{dougan2014ncbi} contains biomedical abstracts annotated with \emph{disease mentions} & \textbf{Sentence}: Twins with AS were identified from the Royal National Hospital for Rheumatic Diseases database.
    \textbf{Entities}: ['AS', 'Rheumatic Diseases']\\ \midrule

    \textbf{MEDM} & \cite{mohan2019medmentions}corpus consists of biomedical abstracts with annotations for \emph{biomedical concepts} that can be found in knowledge bases. & \textbf{Sentence}: A premature electrical impulse from one of four grid corners was utilized to initiate activation.
    \textbf{Entities} : ['premature', 'electrical impulse', 'initiate', 'activation']\\
    \\*
    
    \textbf{PICO} & The EBM-NLP corpus \cite{nye-etal-2018-corpus} contains clinical trial abstracts annotated with \emph{(P)articipants}, \emph{(I)nterventions}, and \emph{(O)utcomes}. & \textbf{Sentence}: Evaluation of lidocaine in human inferior alveolar nerve block. \textbf{Entities} : 'population': ['human inferior alveolar nerve block'], 'intervention': ['lidocaine'], 'outcome': [] \\
    \\*
    
    \textbf{CHIA}  & This dataset  contains text snippets from clinical trial eligibility criteria annotated with entities that can be used to form executable logic statements/queries representing the criteria.\cite{kury2020chia} & \textbf{Sentence}: Use of medications that alter the absorption or metabolism of levothyroxine. 
    \textbf{Entities} : {'Drug' : ['medications', 'levothyroxine'], 'Negation' : ['alter'], 'Observation' : ['absorption of levothyroxine', 'metabolism of levothyroxine'], 'Scope' : ['absorption or metabolism of levothyroxine']}\\
\bottomrule
\end{tabular}
\caption{Overview of all datasets included in our final biomedical NER evaluation testbed.}
\label{tab:datasetexamples}
\end{table*}

\begin{table}[h]
    \centering
    \begin{tabular}{ll}
    \toprule
    \textbf{TUI id} & \textbf{Name of the entity} \\
    \midrule
    T017 & Anatomical Structure \\ 
    T018 & Embryonic Structure \\ 
    T019 & Congenital Abnormality \\ 
    T020 & Acquisalmon Abnormality \\ 
    T021 & Fully Formed Anatomical Structure \\ 
    T024 & Tissue \\ 
    T025 & Cell \\ 
    T026 & Cell Component \\
    T028 & Gene or Genome \\ 
    T032 & Organism Attribute \\ 
    T034 & Laboratory or Test Result \\
    T037 & Injury or Poisoning \\ 
    T038 & Biologic Function \\ 
    T039 & Physiologic Function \\ 
    T040 & Organism Function \\ 
    T041 & Mental Process \\ 
    T045 & Genetic Function \\ 
    T046 & Pathologic Function \\ 
    T047 & Disease or Syndrome \\ 
    T048 & Mental or Behavioral Dysfunction \\ 
    T059 & Laboratory Procedure \\ 
    T060 & Diagnostic Procedure \\ 
    T061 & Therapeutic or Preventive Procedure \\
    T064 & Governmental or Regulatory Activity \\
    T082 & Spatial Concept \\ \bottomrule
    \end{tabular}
    \caption{The final set of categories used for all definition augmentation experiments (Part 1)}
\end{table}

\begin{table}[h]
    \centering
    \begin{tabular}{ll}
    \toprule
    \textbf{TUI id} & \textbf{Name of the entity} \\
    \midrule

    T082 & Spatial Concept \\ 
    T063 & Molecular Biology Research Technique \\ 
     
    T083 & Geographic Area \\ 
    T085 & Molecular Sequence \\ 
    T086 & Nucleotide Sequence \\ 
    T087 & Amino Acid Sequence \\ 
    T088 & Carbohydrate Sequence \\ 
    T089 & Regulation or Law \\ 
    T095 & Self-help or Relief Organization \\ 
    T097 & Professional or Occupational Group \\ 
    T101 & Patient or Disabled Group \\ 
    T121 & Pharmacologic Substance \\ 
    T122 & Biomedical or Dental Material \\
    T123 & Biologically Active Substance \\ 
    T125 & Hormone \\ 
    T126 & Enzyme \\ 
    T127 & Vitamin \\ 
    T129 & Immunologic Factor \\ 
    T131 & Hazardous or Poisonous Substance \\ 
    T169 & Functional Concept \\ 
    T170 & Intellectual Product \\ 
    T191 & Neoplastic Process \\ 
    T192 & Receptor \\ 
    T203 & Drug Delivery Device \\ 
    T204 & Eukaryote \\ \bottomrule
    \end{tabular}
    \caption{The final set of categories used for all definition augmentation experiments (Part 2)}
    \label{tab:tuis}
\end{table}

\clearpage
\section{Implementation Details}
\label{apx:implementation}

We used OpenAI API \footnote{\url{https://platform.openai.com/}}, Anthropic API \footnote{\url{https://console.anthropic.com/}}
and Together API \footnote{\url{https://api.together.xyz/}} to run inference. We use the following settings for all closed source models. Temperature is 0 and max number of tokens for extractions being 256. For generating definitions with GPT-4, we increase the max number of tokens to 4096.
We use the spaCy (en\_core\_web\_sm) library \cite{spacy2} for tagging biomedical entities.

We fine-tune \texttt{Flan-T5-XL} from HuggingFace \cite{wolf2020huggingfaces} library on NVIDIA RTX A6000 GPU. We fine-tune with a learning rate of 1e-3 for 10 epochs. We adapt Low-Rank Adaptation of LLM (LoRA) \cite{hu2021lora} with the following parameters : lora\_alpha: 32, lora\_dropout: 0.05 and SEQ\_2\_SEQ\_LM as the task type.

Output formatting: For datasets with a single entity type (i.e., MEDM and NCBI), we format the outputs as \texttt{entity\_name <sep> entity\_name}; for datasets with multiple types (i.e., CHEM, CDR, PICO and CHIA) we use the format: \texttt{[entity\_name:entity\_type, \ldots, entity\_name:entity\_type]}. 





\begin{table*}[ht]
\centering
\small
\resizebox{!}{3cm}{
\begin{tabular}{llcccccc} 
\toprule
\textbf{Model} & \textbf{Setting} & \textbf{CHEM} & \textbf{CDR} & \textbf{MedM} & \textbf{NCBI} & \textbf{PICO} & \textbf{CHIA}\\
\midrule
\multirow{2}{*}{\textbf{GPT-3.5}}  & 
\textbf{ZS} & 48.61 & 67.65&	43.77&	54.05 &	10.25	& 7.50 \\
& \textbf{SC} & 47.18 & 68.01 & 45.6 & 52.29 & 8.16 & 8.53 \\			

\hline
\multirow{2}{*}{\textbf{Claude 2}}  & 
\textbf{ZS} & 54.28 &	70.07 &	36.98 &	44.17 &	7.26 & 20.12 \\
& \textbf{SC} & 55.43 & 68.75 &	35.55 & 37.28 & 6.9 & 20.17 \\

\hline
\multirow{2}{*}{\textbf{Llama 2}}  & 
\textbf{ZS} & 60.30 & 64.07 & 25.98 &	47.38 &	7.88 & 4.24 \\
& \textbf{SC} & 57.63 & 64.07 & 26.08 & 44.81 &	6.7	& 5.87 \\

\hline
\multirow{2}{*}{\textbf{GPT-4}}  & 
\textbf{ZS} & 62.12 & 70.92 & 47.13 & 54.67 & 7.29 & 16.39 \\
& \textbf{SC} & 63.85 & 71.02 & 46.86 & 56.75 & 7.41 & 16.96 \\

\hline
\end{tabular}
}
\caption{F1 scores of zero-shot (ZS) followed by self-consistency (SC) for all models and datasets. We don't see gain in the performance when prompted without augmenting with the definitions.}

\label{tab:selfconsistency}
\end{table*}




\begin{figure*}[!t]
\centering
  \includegraphics[scale=0.5]
  {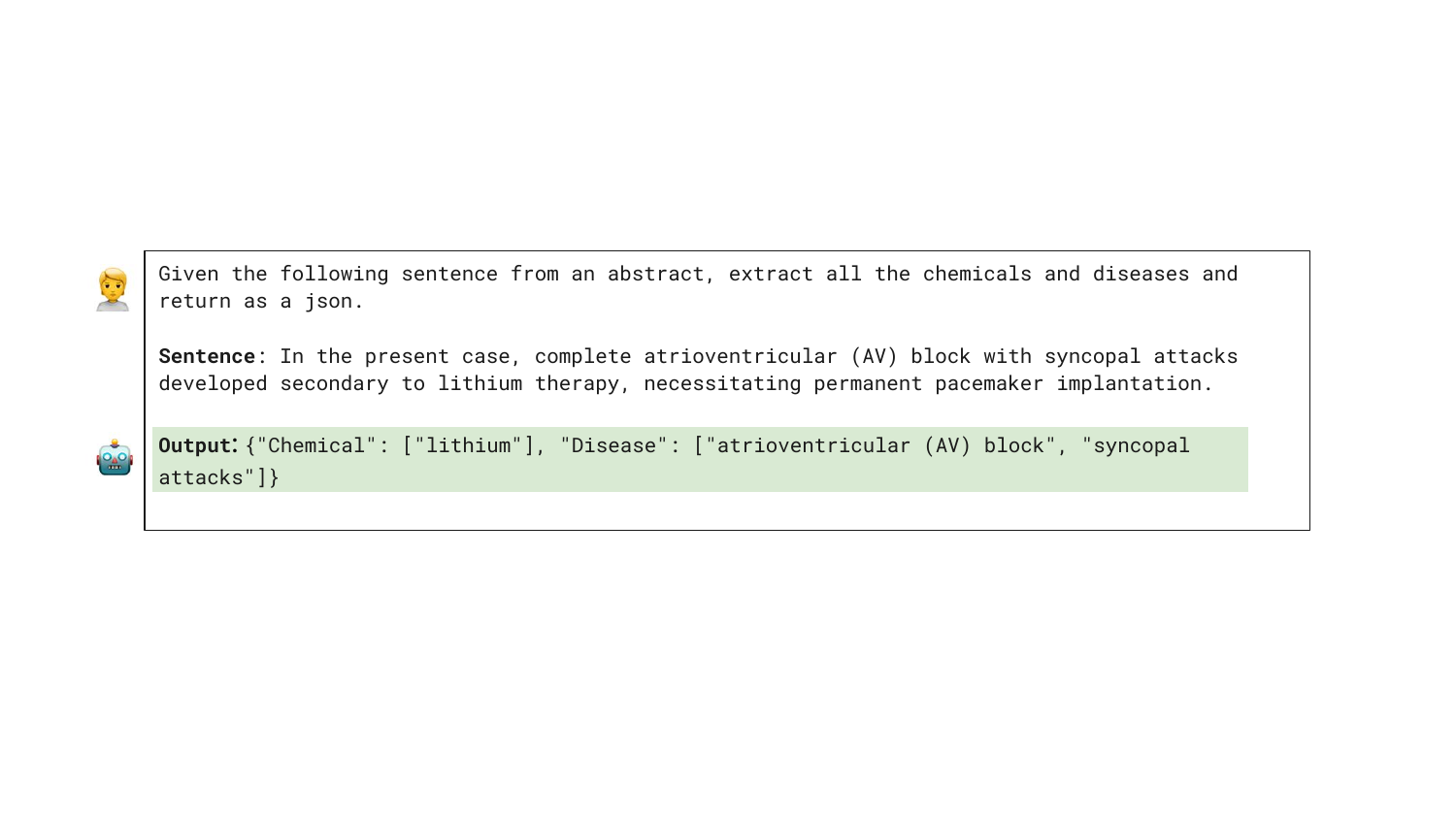}
  \caption{Zero-shot Prompt with \emph{text input} and \emph{JSON output}}
  \label{apx:prompts}
\end{figure*}

\begin{figure*}[!t]
\centering
  \includegraphics[scale=0.5]
  {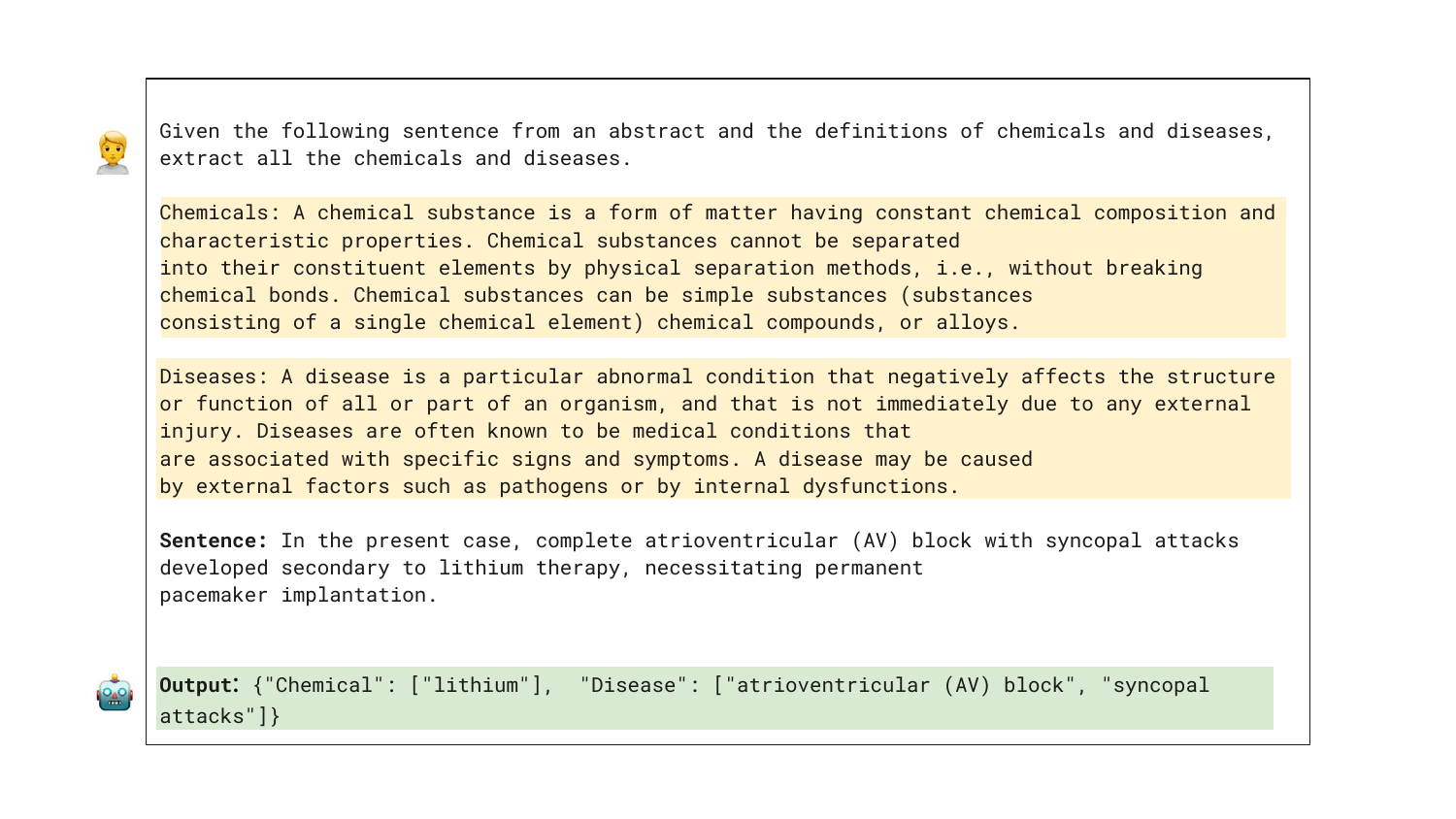}
  \caption{Zero-shot Prompt with \emph{schema def input} and \emph{JSON output}}
\end{figure*}

\begin{figure*}[!t]
\centering
  \includegraphics[scale=0.5]
  {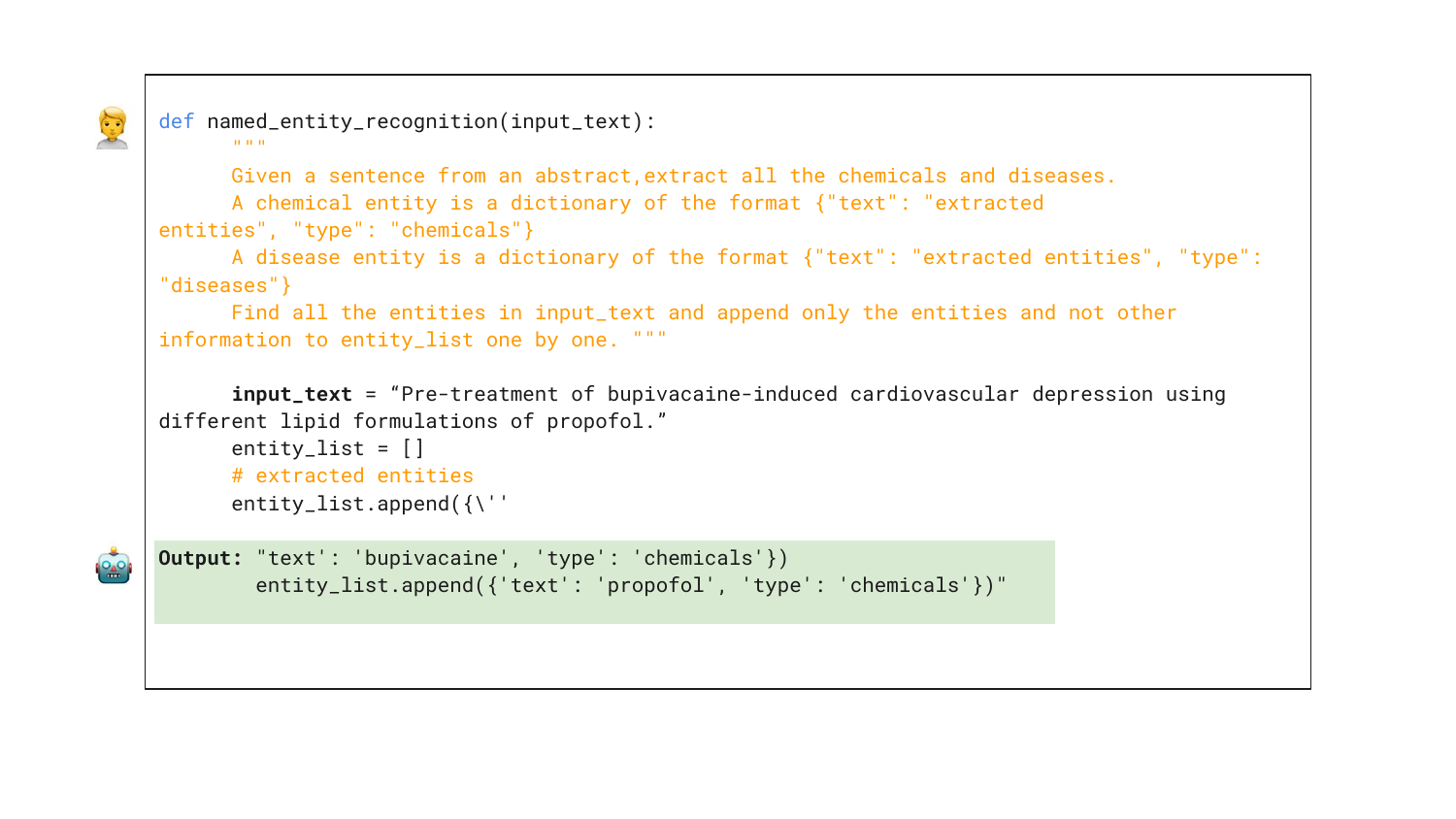}
  \caption{Zero-shot Prompt with \emph{text input} and \emph{code output}}
\end{figure*}

\begin{figure*}[!t]
\centering
  \includegraphics[scale=0.5]
  {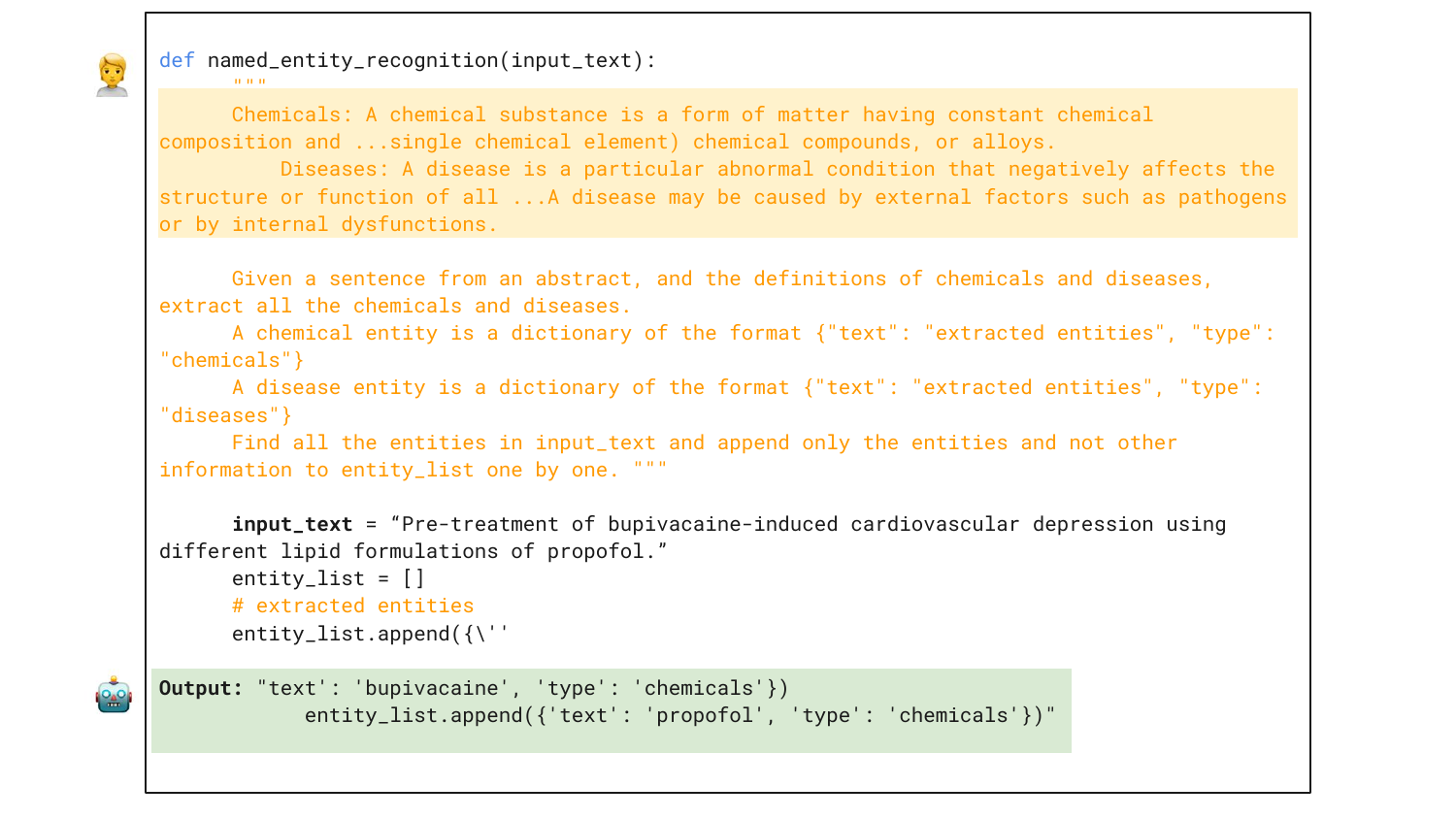}
  \caption{Zero-shot Prompt with \emph{schema def input} and \emph{code output}}
\end{figure*}

\begin{figure*}[!t]
\centering
  \includegraphics[scale=0.5]
  {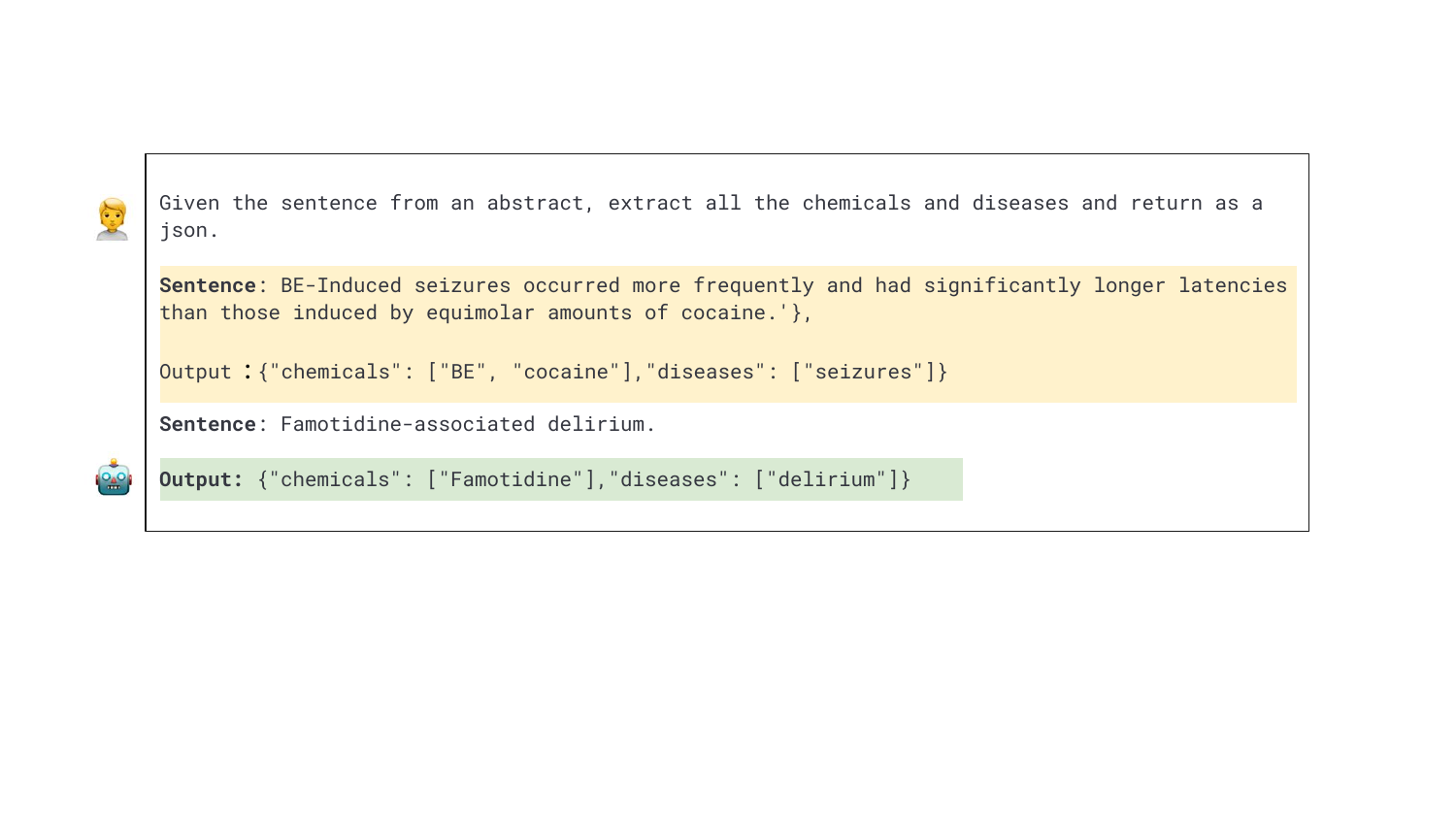}
  \caption{Few-shot Prompt with \emph{text} and \emph{JSON output}}
\end{figure*}

\begin{figure*}[!t]
\centering
  \includegraphics[scale=0.5]
  {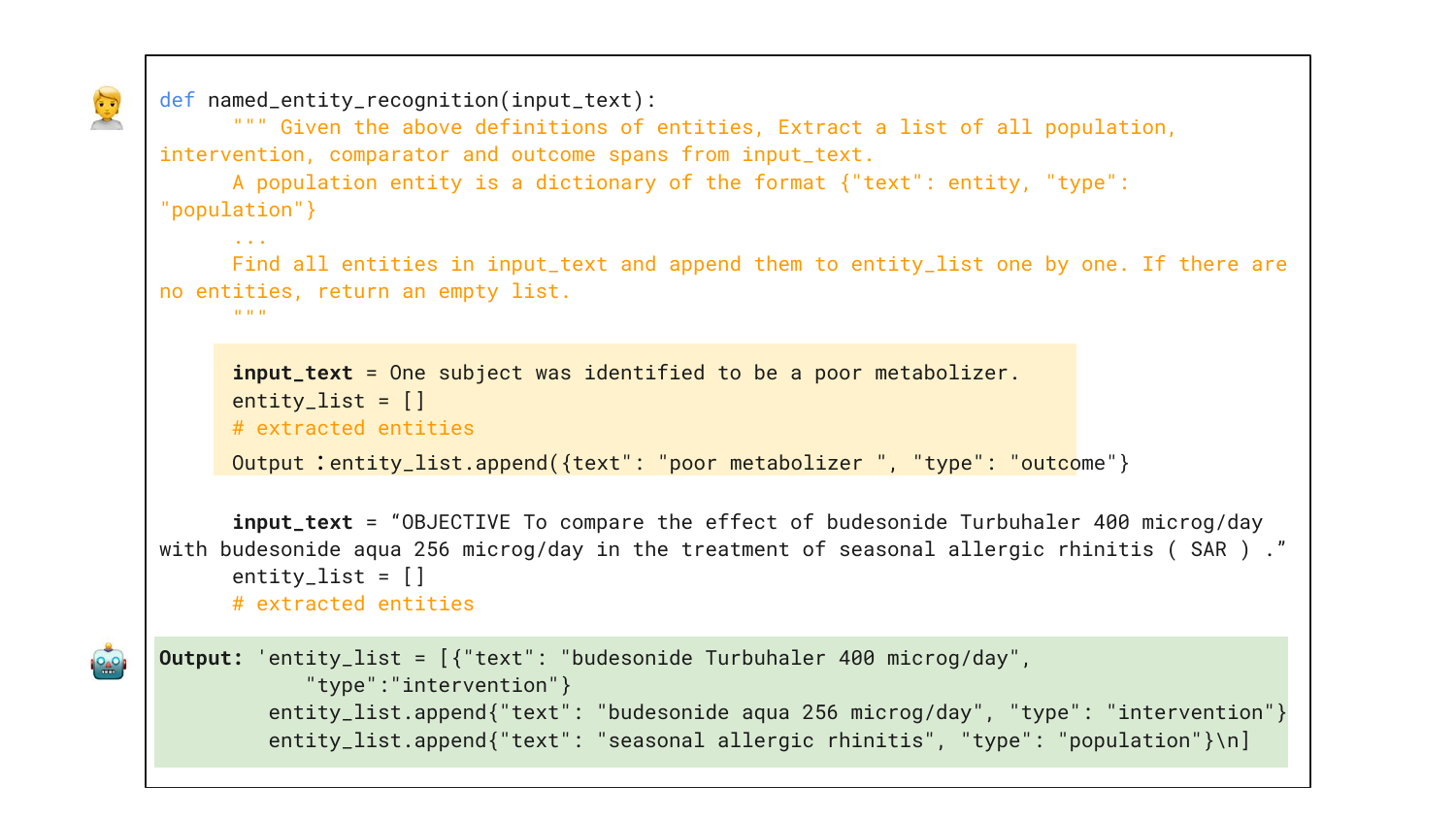}
  \caption{Few-shot Prompt with \emph{text} and \emph{code output}}
\end{figure*}

\begin{figure*}[!t]
\centering
  \includegraphics[scale=0.5]
  {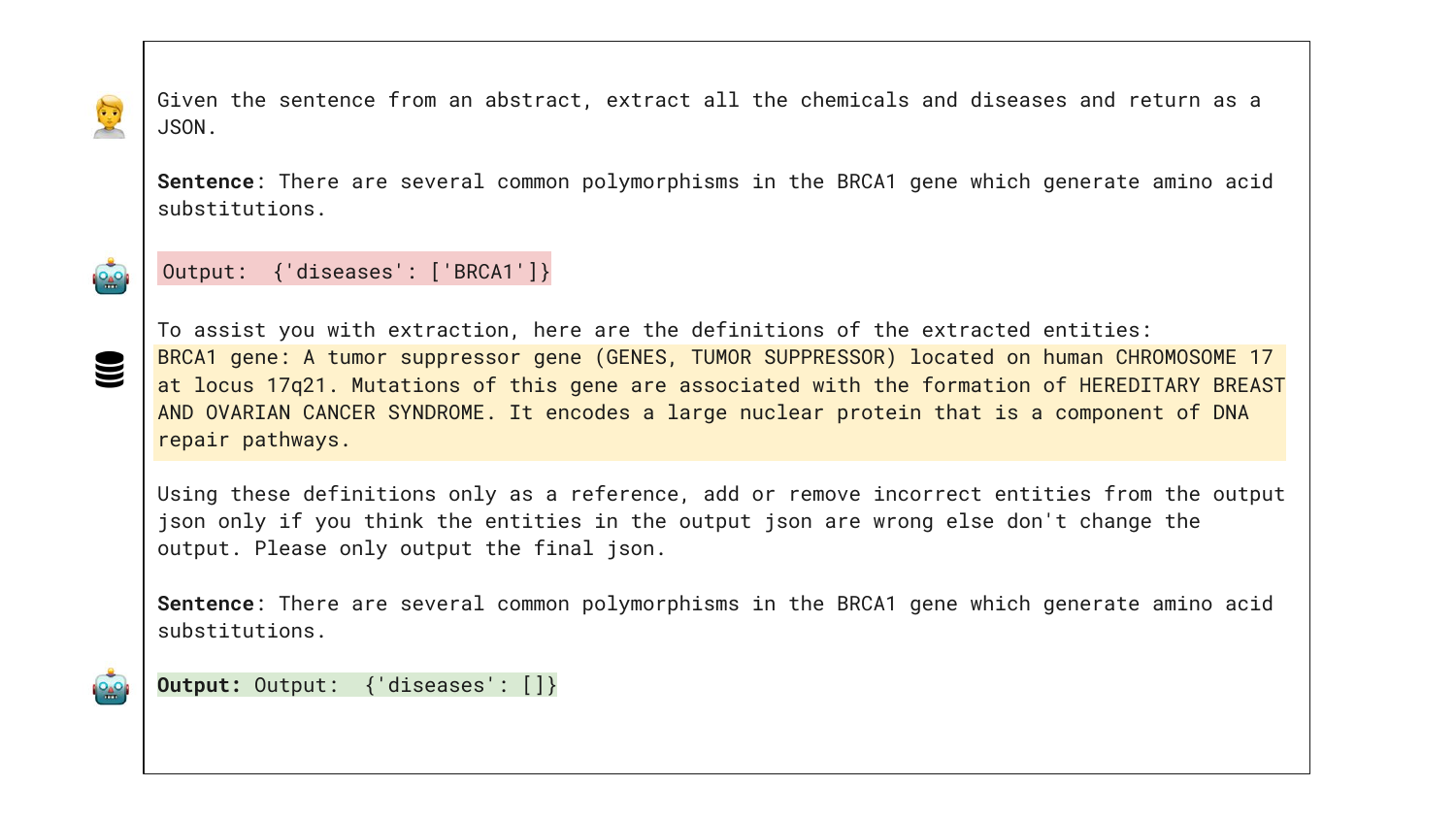}
  \caption{Zero-shot Definition Augmentation with Single Turn}
\end{figure*}

\begin{figure*}[!t]
\centering
  \includegraphics[scale=0.5]
  {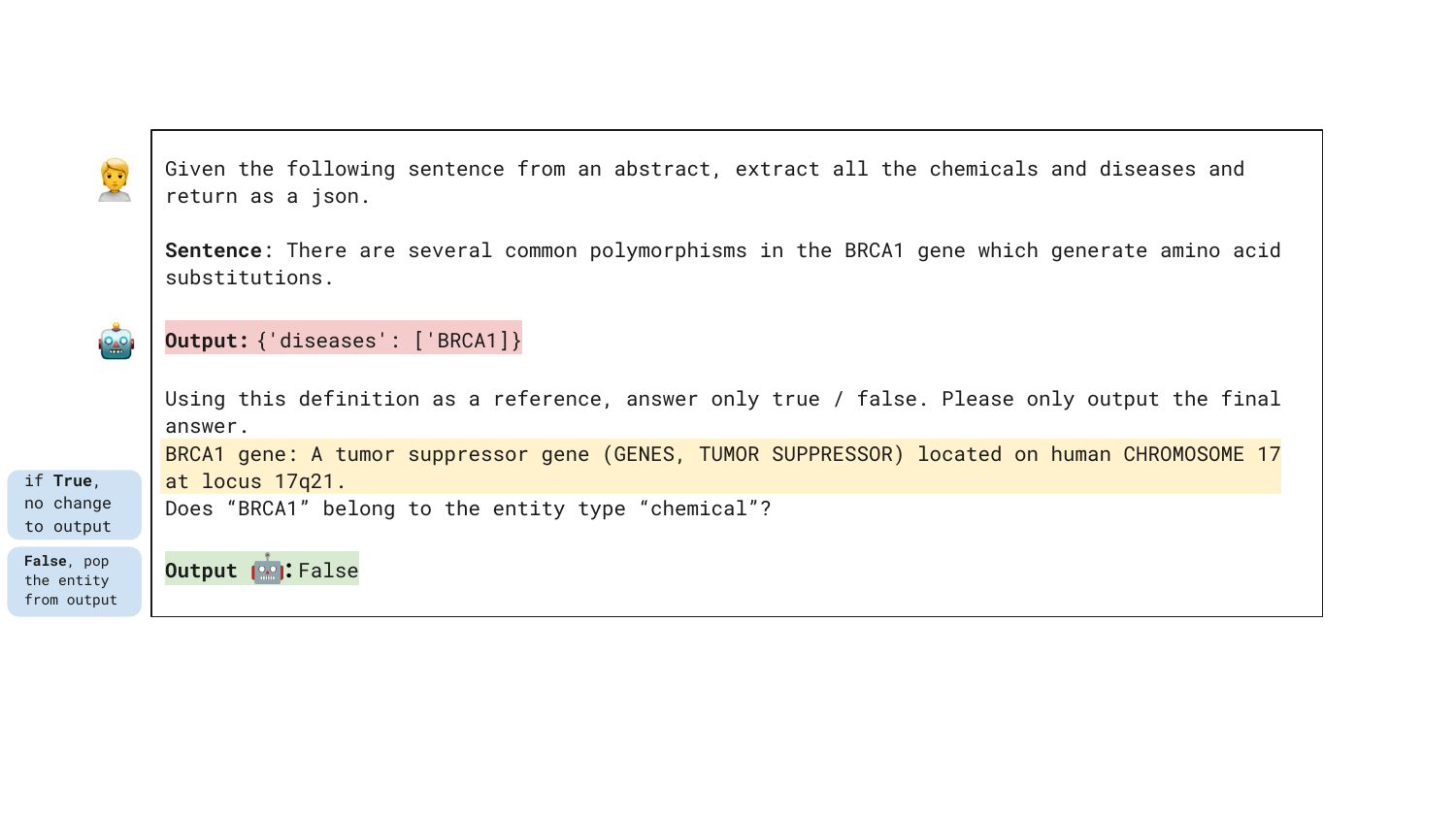}
  \caption{Zero-shot Definition Augmentation with Iterative Prompting with extracted entities}
\end{figure*}

\begin{figure*}[!t]
\centering
  \includegraphics[scale=0.5]
  {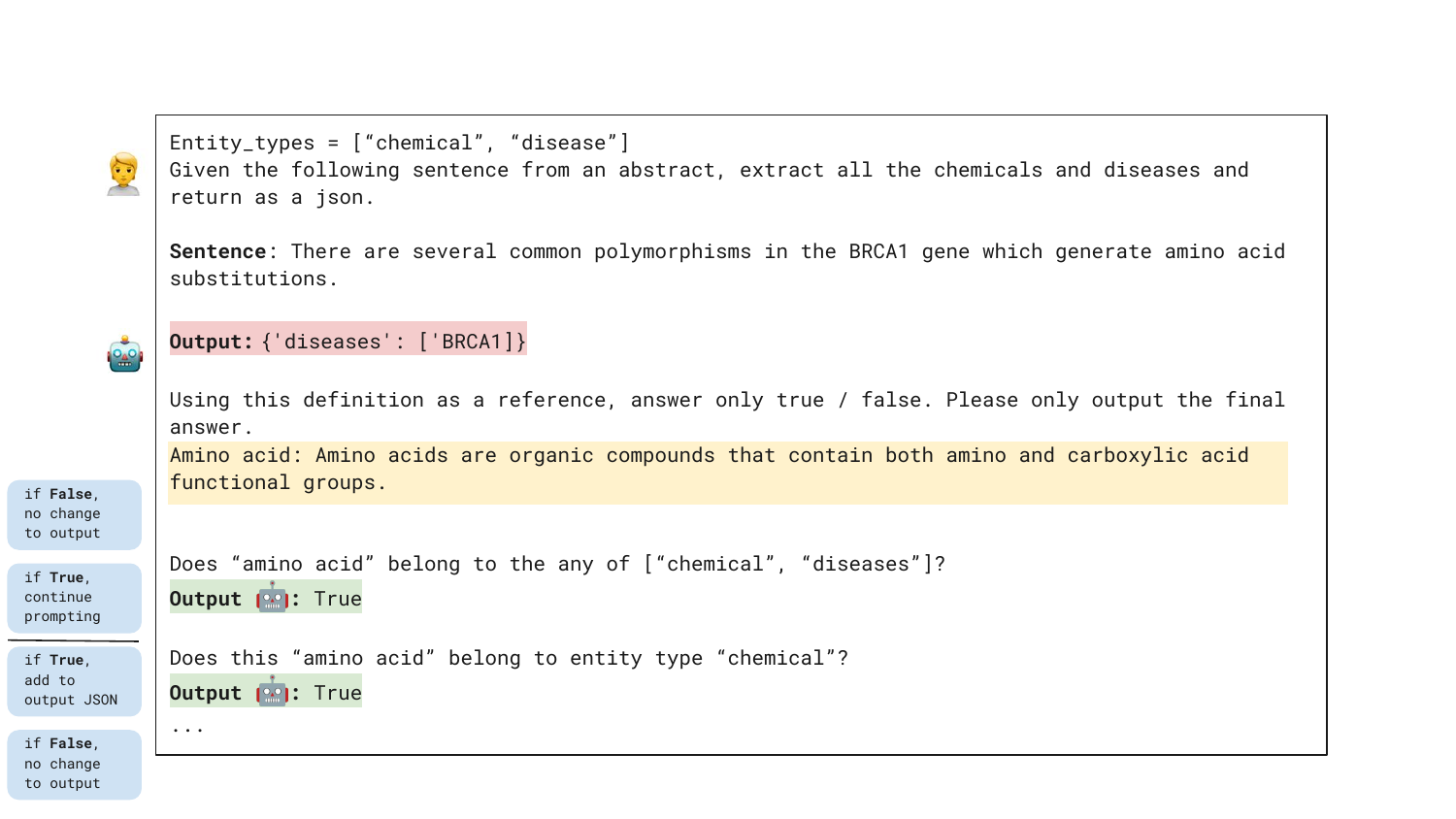}
  \caption{Zero-shot Definition Augmentation with Iterative Prompting with biomedical phrases}
\end{figure*}

\begin{figure*}[!t]
\centering
  \includegraphics[scale=0.5]
  {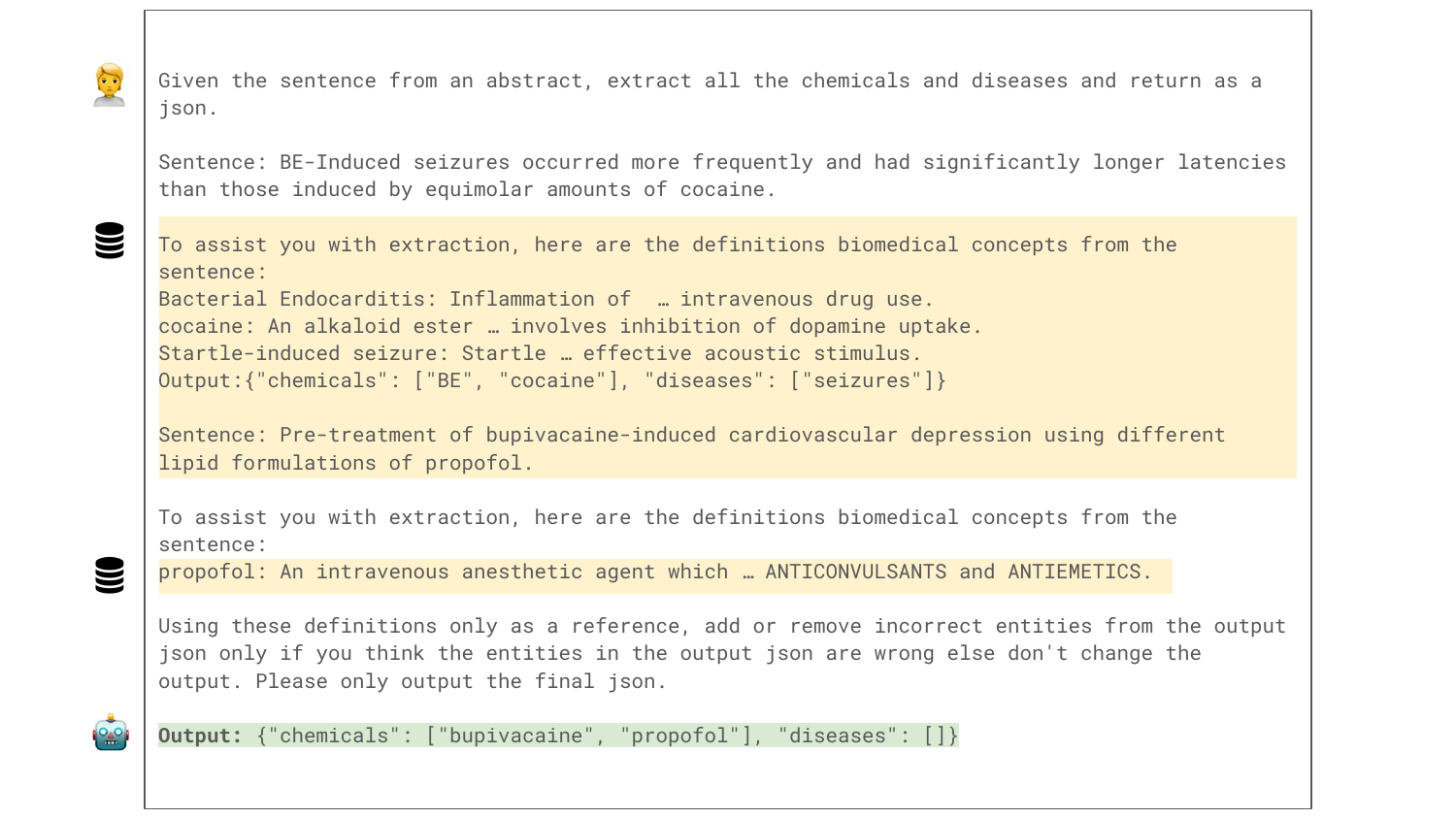}
  \caption{{Few-shot Definition Augmentation with Single Turn}}
\end{figure*}

\end{document}